\documentclass[review]{elsarticle}

\usepackage[utf8]{inputenc}
\usepackage{amsmath, amsfonts, amssymb, amsbsy}
\usepackage{bm}
\usepackage{mathrsfs}

\usepackage{hyperref}

\usepackage[ruled,vlined,linesnumbered]{algorithm2e}
\usepackage[noend]{algpseudocode}
\usepackage{longtable}
\usepackage{pdflscape}
\usepackage{booktabs}
\usepackage{amssymb}
\usepackage{subfig}
\usepackage{multirow}
\usepackage{xcolor}
\usepackage{comment}
\usepackage{graphicx}
\graphicspath{ {./images/} }
\usepackage{caption}
\usepackage{float}

\usepackage{setspace}
\newcommand{\tcolR}{\textcolor{red}}

\begin{document}

\begin{frontmatter}
% \title{DT based forecasting}
\title{Explainable boosted linear regression for time series forecasting} % tentative title

\author[1]{Igor Ilic}
\ead{iilic@ryerson.ca }

\author[2]{Berk Görgülü}
\ead{bgorgulu@mie.utoronto.ca}

\author[1]{Mucahit Cevik\corref{cor1}%
\fnref{fn1}}
\ead{mcevik@ryerson.ca}

\author[3]{Mustafa Gökçe Baydoğan}
\ead{mustafa.baydogan@boun.edu.tr }

\cortext[cor1]{Corresponding author}
\fntext[fn1]{Ryerson University
350 Victoria Street
Toronto, ON, Canada, M5B 2K3}
\address[1]{Ryerson University, Toronto, ON, Canada}
\address[2]{University of Toronto, Toronto, ON, Canada}
\address[3]{Bogazici University, Istanbul, Turkey}

\begin{abstract}
    %Aim
    %Background
    Time series forecasting involves collecting and analyzing past observations to develop a model to extrapolate such observations into the future. Forecasting of future events is important in many fields to support decision making as it contributes to reducing the future uncertainty.
    %Methodology
    We propose explainable boosted linear regression (EBLR) algorithm for time series forecasting, which is an iterative method that starts with a base model, and explains the model's errors through regression trees. At each iteration, the path leading to highest error is added as a new variable to the base model. In this regard, our approach can be considered as an improvement over general time series models since it enables incorporating nonlinear features by residuals explanation. More importantly, use of the single rule that contributes to the error most allows for interpretable results. The proposed approach extends to probabilistic forecasting through generating prediction intervals based on the empirical error distribution.
    %Results
    We conduct a detailed numerical study with EBLR and compare against various other approaches. We observe that EBLR substantially improves the base model performance through extracted features, and provide a comparable performance to other well established approaches.
    %Conclusion
    The interpretability of the model predictions and high predictive accuracy of EBLR makes it a promising method for time series forecasting.
\end{abstract}
\end{frontmatter}

\vspace{-8pt}
%%%%%%%%%%%%%%%%%%%%%%%%%%%%%%%%%%%%%
\section{Introduction}
%%%%%%%%%%%%%%%%%%%%%%%%%%%%%%%%%%%%%
Time series forecasting has important applications in various domains including energy~\citep{deb2017review}, finance~\citep{krollner2010financial} and weather~\citep{baboo2010efficient}. Accurate forecasts provide insights on the trends in the domain, and serve as inputs to decisions involving future events. For instance, in supply chain operations, sales and demand forecasts of the products are essential for inventory control, production planning and workforce scheduling. Accordingly, effective forecasting tools are directly linked to increased profits and reduced costs.

Quantitative forecasting methods are generally divided into two categories: general time series models and regression-based models. General time series models such as exponential smoothing and autoregressive integrated moving average (ARIMA) are typically derived from the statistical information in the historic data. On the other hand, regression models rely on constructing a relation between independent variables (e.g. features such as previous observations) and dependent variables (e.g. target outcomes). There is a wide range of regression approaches used for time series forecasting including linear regression, ensemble methods and neural networks. 

% \tcolR{a discussion about point forecasting vs probabilistic forecasting will be added here!}
While majority of previous studies on time series prediction focus on point forecasts, many applications benefit from having probabilistic/interval forecasts that can provide information on future uncertainty. For instance, in retail businesses, probabilistic forecasts enable generating different strategies for a range of possible outcomes provided by the forecast intervals. A probabilistic forecast typically consists of upper and lower limits, and the corresponding interval can be taken as a confidence interval around the point forecasts. Standard methods such as exponential smoothing and ARIMA generate probabilistic forecasts through simulations or closed form expressions for the target predictive distribution \citep{box2015time}. Recent studies propose deep learning models for probabilistic forecasting that target predicting the parameters (i.e. mean and variance) of the probability distribution for the next time step, and show performance improvements over standard approaches for large datasets consisting of a high number of time series \citep{rangapuram2018deep, salinas2020deepar}.

In predictive modeling, often times the models are evaluated by measuring their prediction performance obtained using a test set based on metrics such as mean absolute error and mean squared error, disregarding the interpretability of the model predictions. However, interpretable models have certain benefits such as creating a trust toward the model by explicit characterization of the factors' contribution to the predictions and providing a better scientific understanding of the model. Value of model interpretability has been acknowledged in recent studies, and lead to new avenues for research \citep{adadi2018peeking, guidotti2018survey}.

Several recent studies on time series forecasting resort to complex deep learning architectures, which typically yield relatively accurate results when the available data is abundant. The drawbacks of such approaches include their computational burden and the black-box nature \citep{rangapuram2018deep, salinas2020deepar}. In this regard, linear models may provide a good trade-off between accuracy and simplicity.
Specifically, linear models with simple mathematical forms are generally preferred for their interpretability and explainability of the model outcomes.

\begin{figure}[!htp]
    \centering
    \includegraphics[scale = 0.58]{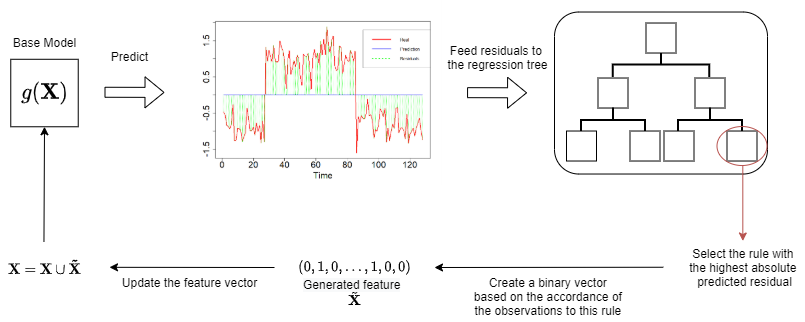}
    \caption{A visual illustration of EBLR.}
    \label{fig:flow}
\end{figure}

This study proposes a time series forecasting method suitable for deterministic and probabilistic forecasting that iteratively improves its predictions through feature generation based on residual exploration. Our model has two stages. In the first stage, a generic forecasting model (i.e. a base learner) is trained to obtain the base forecasts. Second stage explores the residuals (i.e. errors) of the existing model with a regression tree trained on all the available features. This tree identifies the feature spaces resulting in the high error rates as a set of rules. The rule contributing to the error the most is used to generate a new feature to be used by the forecasting model in the first stage. The iterations continue until a certain stopping criterion is met, e.g., certain number of features are generated, the regression tree cannot make a split or forecasting model error cannot be improved further. The idea of learning interaction features based on decision trees is introduced by \cite{friedman2008predictive}. The method proposed in this work extends this idea by generating interaction features based on the unexplored residuals. A visual representation of the proposed method is provided in Figure \ref{fig:flow}.

The proposed method shares similar ideas with gradient boosting regression (GBR) trees. Specifically, GBR trees fit a decision tree on the residuals obtained from the fitted prediction model (or base learner) and aim to improve the prediction model by adding new decision trees to update the base prediction. At each iteration, predicted residuals are used to update the base prediction after they are multiplied with a learning rate. In other words, GBR trees use all the rules to update the prediction with a fixed learning rate. On the other hand, our approach uses the residuals to determine the highest source of errors and creates a single and interpretable feature to the fitted model to improve its predictive performance at each iteration. Compared to the use of a fixed learning rate, the base learner in the first stage is retrained to find an appropriate weight to the introduced feature. Unlike generic boosting algorithms, our method does not sacrifice interpretability while improving the prediction performance. Moreover, our method generates a fewer but more meaningful features, which leads to a more memory efficient method.

The basic time series models and other linear models rely on trends, and seasonality, and leave unexplained components as noise. Due to exponential number of potential interactions, kernel-based methods (\cite{muller1997predicting}) are introduced to capture these interactions and nonlinearity without explicitly introducing the features to the model. However, these kernels cause model to lose its interpretability. The proposed approach identifies interactions between features through a regression tree in the second stage and explicitly adds this information to the model. Use of the single set of rules that contributes to the error most not only improves the predictive performance, but also allows for interpretable results. For instance, for retail sales forecasting, if there is an implicit interaction between holiday and promotion day variables that creates a higher effect than their individual effects, a new (nonlinear) variable is included in the fitted model. This variable implies that the promotion has a higher effect on the sales when applied on a holiday. In a similar manner, our approach is able to capture many polynomial interaction terms and incorporate those to the fitted model depending on the choice of the base learner (i.e. linear regression).

After generating point forecasts, the proposed model extends to probabilistic forecasting. It generates probabilistic forecasts based on the empirical error distribution which is representative of the underlying real distribution. Prediction intervals are generated by extracting quantiles through bootstrapping the residuals of the time series~\citep{davison1997bootstrap}.

The rest of this paper is organized as follows. In Section \ref{sec:relatedWorks} we provide a review of the relevant works. Section \ref{sec:eblr} introduces EBLR and its extensions with necessary backgrounds. Section \ref{sec:num_experiments} presents the numerical experiments and the conclusion is provided in Section \ref{sec:conclusion}.
% \begin{itemize}
%     \item[] INTRO TO-DO LIST:
%     \item Explain the importance of interpretability [DONE?].
%     \item Motivate usage of linear models focusing on their interpretability [DONE?].
%     \item List linear time series prediction methods and associated problems (with citations) [DONE in Related works?].
%     \item Introduce non-linear complex methods by mentioning that they improve the performance by sacrificing from the interpretability [DONE?].
%     \item Propose nonlinear feature generation as a midway approach[??].
%     \item Too many potential features including polynomial and interaction terms [DONE?].
%     \item Propose our approach [DONE?].
%     \item Talk about similarities and differences from the boosting methods [DONE?].
% \end{itemize}

%%%%%%%%%%%%%%%%%%%%%%%%%%%%%%%%%%%%%
\section{Related Works}\label{sec:relatedWorks}
%%%%%%%%%%%%%%%%%%%%%%%%%%%%%%%%%%%%%

% general review on time series regression including deep learning models
Time series forecasting has been well studied in the literature. While the earlier studies focused on linear statistical forecasting methods such as exponential smoothing and ARIMA, nonlinear models including neural networks has been shown to perform better for various forecasting tasks \citep{alon2001forecasting, chu2003comparative}. Decision trees and their ensembles such as random forests and gradient boosted trees were also frequently used for time series forecasting especially due to interpretability of the model predictions~\citep{galicia2019multi, taieb2014gradient}. Recent studies particularly focused on deep learning and metalearning approaches, reporting substantial performance improvements for the time series forecasting task over the linear models as well as standard supervised learning approaches \citep{ma2020retail, rangapuram2018deep, salinas2020deepar}. While these models show significant promise in generating highly accurate results especially when data is abundant, they are typically regarded as black-box models, which are not deemed as interpretable/explainable. We refer the reader to \citet{parmezan2019evaluation}'s study for a detailed overview of statistical and machine learning models for time series forecasting.

% hybrid models
Several hybrid approaches have been considered to incorporate nonlinear relations between input and output variables. \citet{zhang2003time} assumed that each time series can be represented as a combination of linear and nonlinear components and developed a hybrid ARIMA and artificial neural network (ANN) model for forecasting. Predictions in the model were obtained as a combination of ARIMA's forecast for the linear component and ANN's forecast for the nonlinear component. \citet{khashei2012methodology} investigated the performances of hybrid forecasting models by comparing generalized hybrid ARIMA/ANN model, \citet{zhang2003time}'s hybrid ARIMA/ANN model and ANN($p,d,q$) models. Their analysis with three different datasets showed that generalized hybrid ARIMA/ANN model, which aimed to find linear relations using ARIMA in the first stage and nonlinear relations using ANNs in the second stage, performed best among the three approaches. \citet{aladag2009forecasting} replaced the feed forward neural network in \citet{zhang2003time}'s model with a recurrent neural network (RNN), which lead to improvements in forecasting accuracy. 
\citet{taskaya2005comparative} performed comparative analysis on ARIMA and ANN hybrids using nine different datasets, showing that components of the hybrid models outperformed their hybrid counterparts in five of the nine instances. They concluded that careful selection of the models to be combined is important for the success of the hybrid models. \citet{arunraj2015hybrid} proposed a hybrid model with seasonal ARIMA (SARIMA) and quantile regression where the latter was used for forecasting the quantiles rather than individual data points. Various other studies considered hybrid models constructed from variants of ARIMA models such as SARIMA and SARIMAX \citep{cools2009investigating, cornelsen2012impact}.

% a review on gbr/rf??

% probabilistic forecasting review
While the common approach for forecasting is the prediction of the expected value of a target value, understanding the uncertainty of a model's predictions can be important in different areas such as macroeconomics and financial risk management. Accordingly, many studies on forecasting focus on modeling uncertainties that lead to probabilistic forecasts. A commonly used approach is to use quantile regression~\citep{liu2015probabilistic}, while some other studies consider ensemble of learned models to generate probabilistic forecasts~\citep{ahmed2016ensemble}. Recent studies focus on deep neural networks to generate mean and variance parameters of the predictive distributions. Specifically, \citet{salinas2020deepar} propose DeepAR model, which is an autoregressive recurrent network-based global model that consider observations from different training time series to build a single probabilistic forecasting model. \citet{rangapuram2018deep} combine state space models with deep learning by parametrizing a linear state space model using an RNN. \citet{wang2019deep} integrate global deep neural network backbone with local probabilistic graphical models where global structure extracts complex nonlinear patterns and local structure captures individual random effects. Combining different probabilistic forecasting methods, \citet{alexandrov2019gluonts} provide an extensive Python library of probabilistic time series models.

% models exploring residuals
Few other studies in the literature focused on building a forecasting model through residual exploration. \citet{aburto2007improved} considered a combined ARIMA and neural network model where the ARIMA model was used to model the original time series and the neural network was used to predict possible forecasting errors. The resulting forecast was taken as the summation of the predicted values by these two models. \citet{gurali2016multi} proposed a two-stage information sharing model for multi-period retail sales forecasting problem. In their model, the first stage estimated various variables such as calendar, seasonality and promotions through a regression analysis, and second stage extrapolated the residual time series. The resulting prediction was obtained by combining the forecasts from the first stage with the extrapolated residuals from the second stage.

%%%%%%%%%%%%%%%%%%%%%%%%%%%%%%%%%%%%%

% \begin{itemize}
%     \item Decision tree parameters: Mainly complexity parameter
%     \item See chapter 4 - https://cran.r-project.org/web/packages/rpart/vignettes/longintro.pdf
% \end{itemize}

%%%%%%%%%%%%%%%%%%%%%%%%%%%%%%%%%%%%%
\section{EBLR}\label{sec:eblr}
%%%%%%%%%%%%%%%%%%%%%%%%%%%%%%%%%%%%%
This section introduces our framework for time-series forecasting and non-linear feature generation called Explainable Boosted Linear Regression (EBLR). Our framework consists of two stages that are applied recursively: model training and feature generation. The first stage is intuitive and utilizes well-known linear models such as linear regression, least absolute shrinkage and selection operator (LASSO) regression, or ARIMA. The second stage generates non-linear features based on regression tree transformation. Below, we first provide necessary background on the regression trees and tree-based representation, then introduce EBLR and its extensions.

\subsection{Regression Trees and Tree-based Representation}
Regression trees \citep{breiman1984classification} are tree structures that recursively partitions the observation space based on some rules. These rules are greedily generated by evaluating all possible splits in the data and selecting the one that provides the highest mean squared-error (MSE) reduction in the children nodes.

Each terminal node in a regression tree can be represented by a set of rules. Due to the nature of the split formation in the trees, each terminal node refers to a hyperrectangle in the feature space (assuming that all features are numerical without loss of generality). From probabilistic view, regression trees model mixture of Gaussian distributions \citep{crim2012decforest}. Feature representations based on the tree structures are shown to provide successful results in different domains and they are sometimes referred to as hashing \citep{lin2014fast,zhou2019deep}. Similarly, each observation is represented by a binary vector based on its presence or absence in a terminal node. For instance, an observation residing in the 3$rd$ node of a regression tree with 5 terminal nodes is represented as $(0,0,1,0,0)$. This representation implicitly encodes the feature space based on the distribution of the target variable \citep{breiman1984classification}. 

\subsection{Methodology}\label{sec:methodology}
Consider a time series dataset containing $N$ time series of length $T$. Let $y^{(i)}_t$ represent the observation at time $t$ of the $i^{th}$ time series and, assume that there is a $(1 \times F)$ feature vector associated with each observation $y^{(i)}_t$, represented by $X^{(i)}_t$. Moreover, $\mathbf{y}$ and $\mathbf{X}$ respectively represent the vector of all observations of size ($N \times 1$) and the matrix that contains complete feature space of size ($N \times F$).

In the first phase, an initial feature set of size $f \in \{1,\dots F\}$ is selected. This feature set can contain a single feature of time information $(t)$ or a collection of features and is represented by an ($N \times f$) matrix called $\mathbf{X'}$. Then, a linear base learner $g(\bm{x})= \alpha + \beta^T \bm{x}$ that maps $\mathbf{X'}$ to a $\mathbf{\hat{y}} \in \mathbb{R}^n$ vector of size $(N \times 1)$ is chosen. The base learner is trained on ($\mathbf{X'}, \mathbf{y}$) pair to obtain the base model parameters $(\beta)$ by minimizing a targeted loss function. Based on the prediction vector $\mathbf{\hat{y}}$ obtained from $g(\mathbf{X'})$, the residuals are calculated and represented as $\mathbf{e} = \mathbf{y} - \mathbf{\hat{y}}$. Note that due to the linear nature of the base learner, residuals potentially contain additional information that is related to the nonlinear patterns in the feature vectors and/or interactions.

In the second phase, the residuals obtained from the base learner are examined to model the unexplored components. Here a regression tree that predicts $\mathbf{e}$ using $\mathbf{X}$ is trained. Each observation $e_t^{i} \in \mathbf{e}$ resides in a single terminal node of the regression tree. Regression trees are constructed to minimize MSE of the target values that end up in each of the terminal nodes. Therefore, when they are trained on the residuals, they essentially aim to group the errors from the same sources. That is, it discovers the undiscovered, potentially nonlinear features that explains a proportion of the errors. Each terminal node in the regression tree represents an error source. Here the terminal nodes are selected such that absolute value of the target means (i.e. residuals) in the terminal nodes is the largest, i.e., terminal node with the largest error source. 

The selected terminal node can be represented with a binary vector $\mathbf{\Tilde{X}} \in \{0,1\}^N$ such that $\mathbf{\Tilde{X}}_i = 1$ if the observation $i$ resides in the selected terminal node and $0$ otherwise. Once this vector is generated, it is added to the current feature vector $\mathbf{X'}$, which is updated by setting $\mathbf{X'} = \mathbf{X'} \cup \mathbf{\Tilde{X}}$. This process is repeated until a stopping condition is met.

\subsection{Parameters}
There are three sets of parameters involved in EBLR: (1) hyperparameters of the base learner, (2) hyperparameters of the regression trees and (3) stopping condition related parameters. Firstly, hyperparameters for the selected base learner can be specified based on the prior knowledge and/or through hyperparameter tuning. Since we focus on linear models such as simple linear regression and LASSO regression, only hyperparameter that should be specified is LASSO penalty which can be easily determined by cross-validation.

Secondly, EBLR requires specification of the decision tree hyperparameters, namely, tree-depth, complexity parameter or minimum observation in terminal nodes. Determining the depth of the tree is of great importance for our method because it directly determines the degree of nonlinearity, i.e., degree of complexity, for the generated features. There are two main approaches that could be used for determining the depth: pre-pruning (determining the tree depth before construction) and post-pruning (pruning the leaf nodes after construction). For our purpose, utilization of pre-pruning is challenging since the ``right amount of complexity'' required for each generated feature is not known apriori to tree construction, therefore, it might need cross-validation by re-constructing the regression tree for many times which might introduce an additional complexity. Unlike pre-pruning, post-pruning based on complexity parameters provide a highly efficient pruning strategy that allows generating features of various complexities in each iteration without re-constructing the regression trees. Post-pruning requires an initial complexity parameter, $\eta$, to be specified. Our preliminary analysis show that setting $\eta$ to a small value would be enough for our method to perform well. 

Lastly, the parameters are specified for the stopping criteria. In this work, the number of features to be generated $(F^{\max})$ is used as the stopping criteria which essentially implies the number of iterations that EBLR runs. It is important to note that there is a trade-off between the base model selection and stopping criteria. If the selected base model is a primitive learner such as simple linear regression, then $F^{\max}$ significantly effects the degree of over-fitting that could be faced with and it should be carefully selected. Whereas, if our base learner is a penalized method such as LASSO regression, $F^{\max}$ could be selected as a very large integer and coefficient of penalization can be tuned to eliminate the features that cause over-fitting. A pseudocode of EBLR is provided in Algorithm \ref{pseudocode}.

% \begin{itemize}
%     %\item Parameters: Number of features + initial complexity parameter
%     %\item Stops learning when number of features is reached or pruned decision tree results in just the root node (all impurity at root node)
%     %\item Talk about importance for post-pruning (rpart) vs pre-pruning trees (sklearn). Specifically post-pruning allows for more generalized features. 
%     %\item Our model prune by using the complexity parameter from the minimum cross validation error
%     \item R.Hyndman Prediction intervals from bootstrapped residuals: https://otexts.com/fpp3/prediction-intervals.html
% \end{itemize}

%%%%%%%%%%%% Here is the algorithm I've made, using algorithm2e - Igor (adjust whatever you need, or you can move it to another package too)
\begin{algorithm}[!h]
    \caption{Pseudocode of EBLR}
    \label{pseudocode}
    \SetAlgoLined
    \SetKwInOut{Input}{Input}
    \SetKwInOut{Output}{Output}
    \Input{Input Dataset $\mathcal{D}$, Number of Features $F^{\max}$, Initial Complexity Parameter $\eta$}
    \Output{Trained model $g$}
    Construct initial feature matrix $\mathbf{X'}$.\;
   % $y \gets \textbf{ExtractTarget}(\mathcal{D})$ // Select target variable\;
   % $\Theta \gets \emptyset$ // Initialize learned features\
    \For{$i = 1\hdots F^{\max}$}{
        Train the base model g on ($\mathbf{X'},\mathbf{y}$)\;
        Calculate the residuals $\mathbf{e} = \mathbf{y} - \mathbf{\hat{y}}$ \;
        Train a regression tree on the residuals based on $\eta$ and using all features $\mathbf{X}$ \;
        Apply post-pruning to the regression tree.\;
        Extract a feature $\mathbf{\Tilde{X}}$ from the terminal nodes with the largest absolute error \;
        Update $\mathbf{X'} = \mathbf{X'} \cup \mathbf{\Tilde{X}}$\;
    }
    Update feature space to include all raw features $\mathbf{X'} = \mathbf{X'} \cup \mathbf{X}$ \;
    Train the base model g on ($\mathbf{X'},\mathbf{y}$)\;
    return $g$\;
\end{algorithm}
%%%%%%%%%%%% %%%%%%%%%%%% 

\subsection{Complexity Analysis}
We conduct a worst-case theoretical complexity analysis for EBLR. The complexity of the method is dictated by the feature generation phase due to simple and linear nature of the base regressors. We take $B$ as the complexity of the base regressor and focus on the feature generation phase.  

Consider a decision tree of depth $d$ constructed on a time series database of size $N \times T$. Then, in the vertical format it corresponds to $NT$ number of observations. Since each observation requires $d$ number of comparisons, the worst-case complexity of the feature generation is $O(NTd)$. Then, the complexity of each iteration becomes $O(NTd +B)$. This process is repeated for $F^{\max}$ times, which yields $O(F^{\max}(NTd+B))$ complexity. Note that the complexity of EBLR is similar to both gradient boosting regression (GBR) and random forest (RF) which is $O(F^{\max}NTd)$ assuming $F^{\max}$ number of iterations.

\subsection{Illustration}\label{sec:illustration}
In order to provide a better understanding, we illustrate EBLR on a simple example. Consider a time series coming from the following model, initialized with $y_0 = y_1 = 0$:
\begin{align*}
    y_t &= -0.4 y_{t-1} + 0.5 y_{t-2} + 5500* isWeekend*isPromotion\\
    &+ 1500*isPromotion + 3000*isWeekend + \mathcal{N}(5000, 150).
\end{align*}

\noindent In this example, in addition to the auto-regressive terms, there are two factors affecting the sales: day of the week and promotion. Also, note that the interaction of these features have heterogeneous effects and are also important. For example, there is a significant difference in the effect of promotion on weekdays and weekends. 

\begin{figure}[!h]
    \centering
    \subfloat[Iteration 0 \label{fig:learnprog0}]{\includegraphics[width=0.5\textwidth]{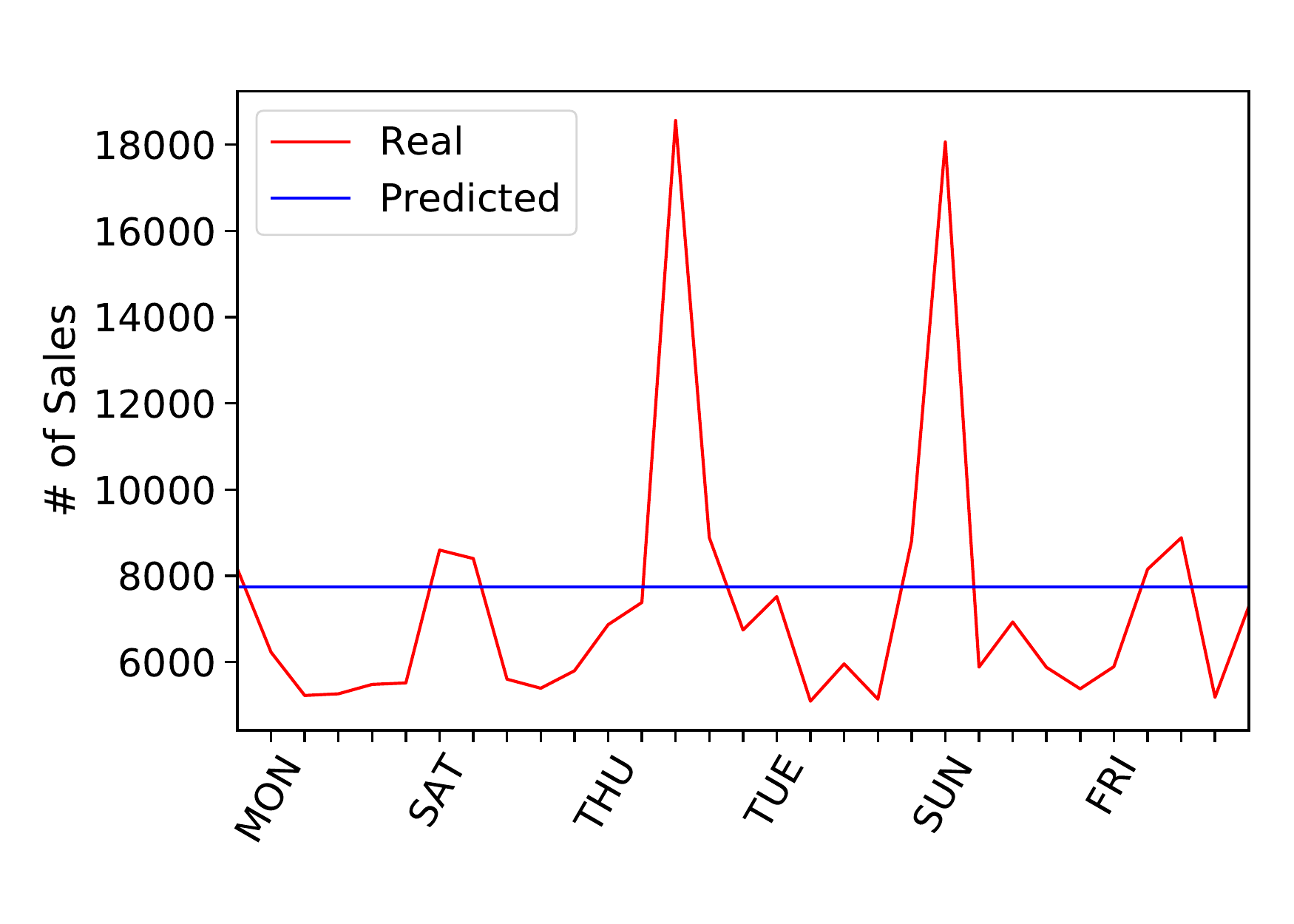}}
    \hfill 
    \subfloat[Iteration 1 \label{fig:learnprog1}]{\includegraphics[width=0.5\textwidth]{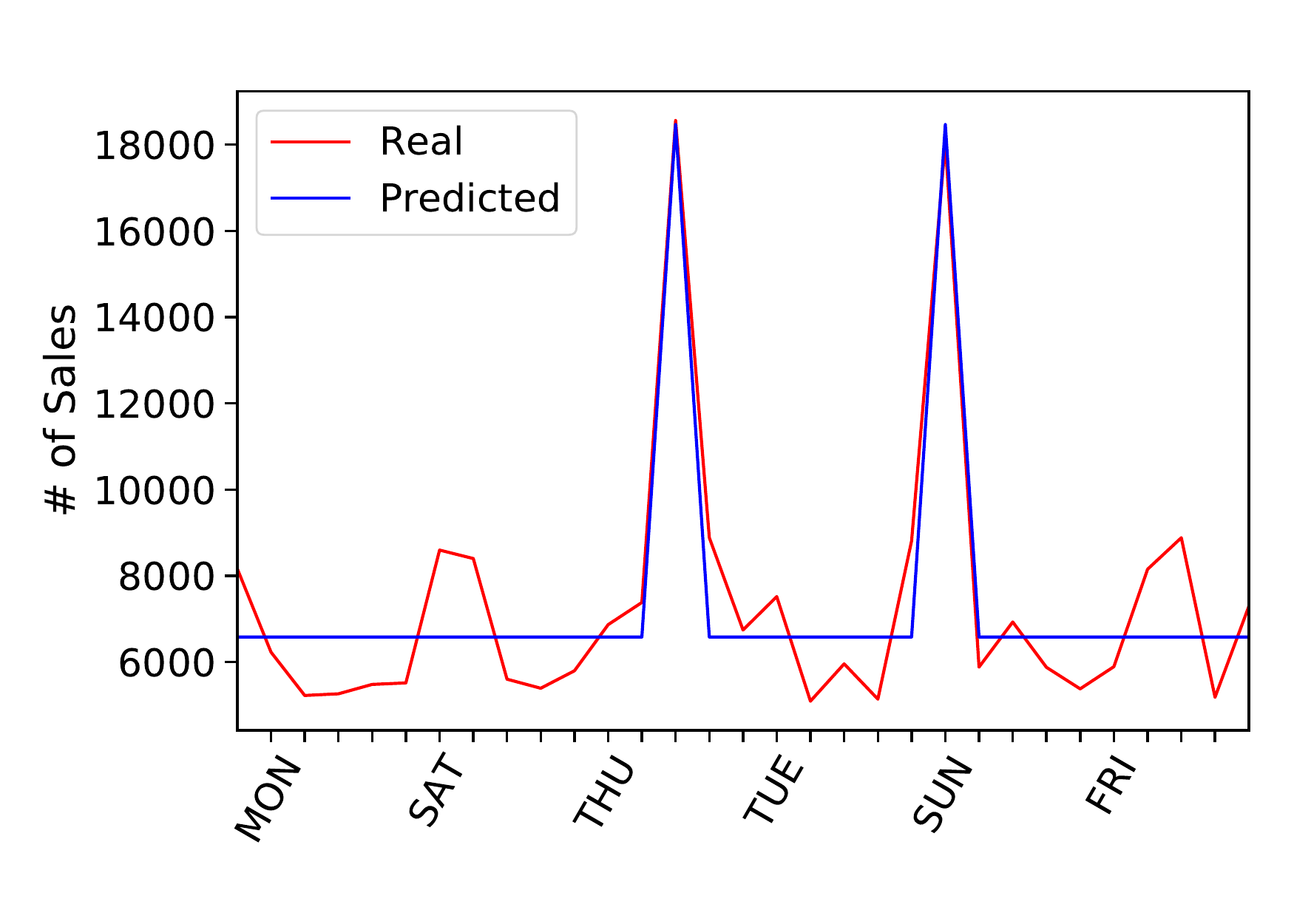}}\\
    \hfill
    \subfloat[Iteration 2 \label{fig:learnprog2}]{\includegraphics[width=0.5\textwidth]{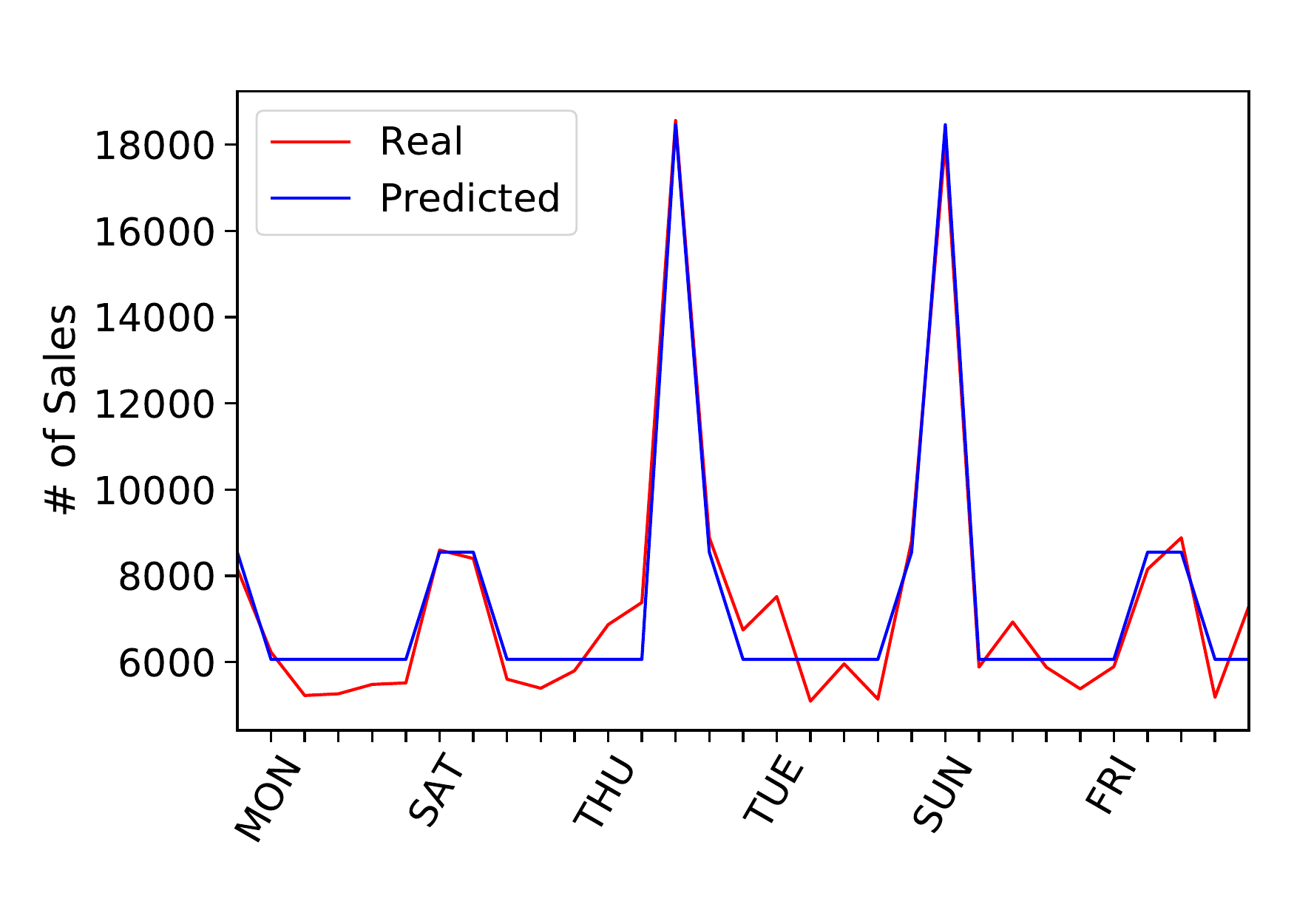}}
    \subfloat[Iteration 3 \label{fig:learnprog3}]{\includegraphics[width=0.5\textwidth]{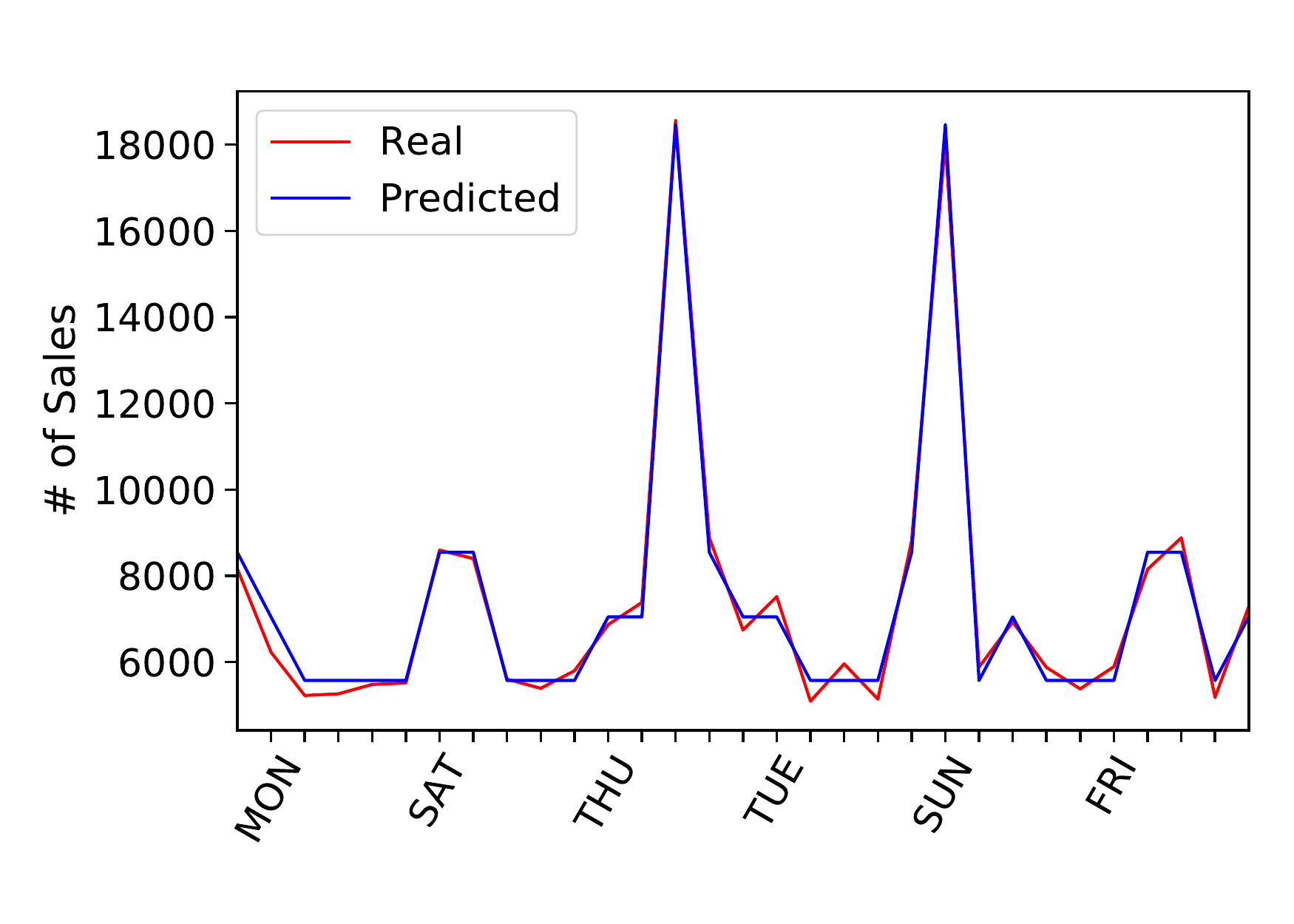}}\\
    \hfill
    \subfloat[Iteration 4 \label{fig:learnprog4}]{\includegraphics[width=0.5\textwidth]{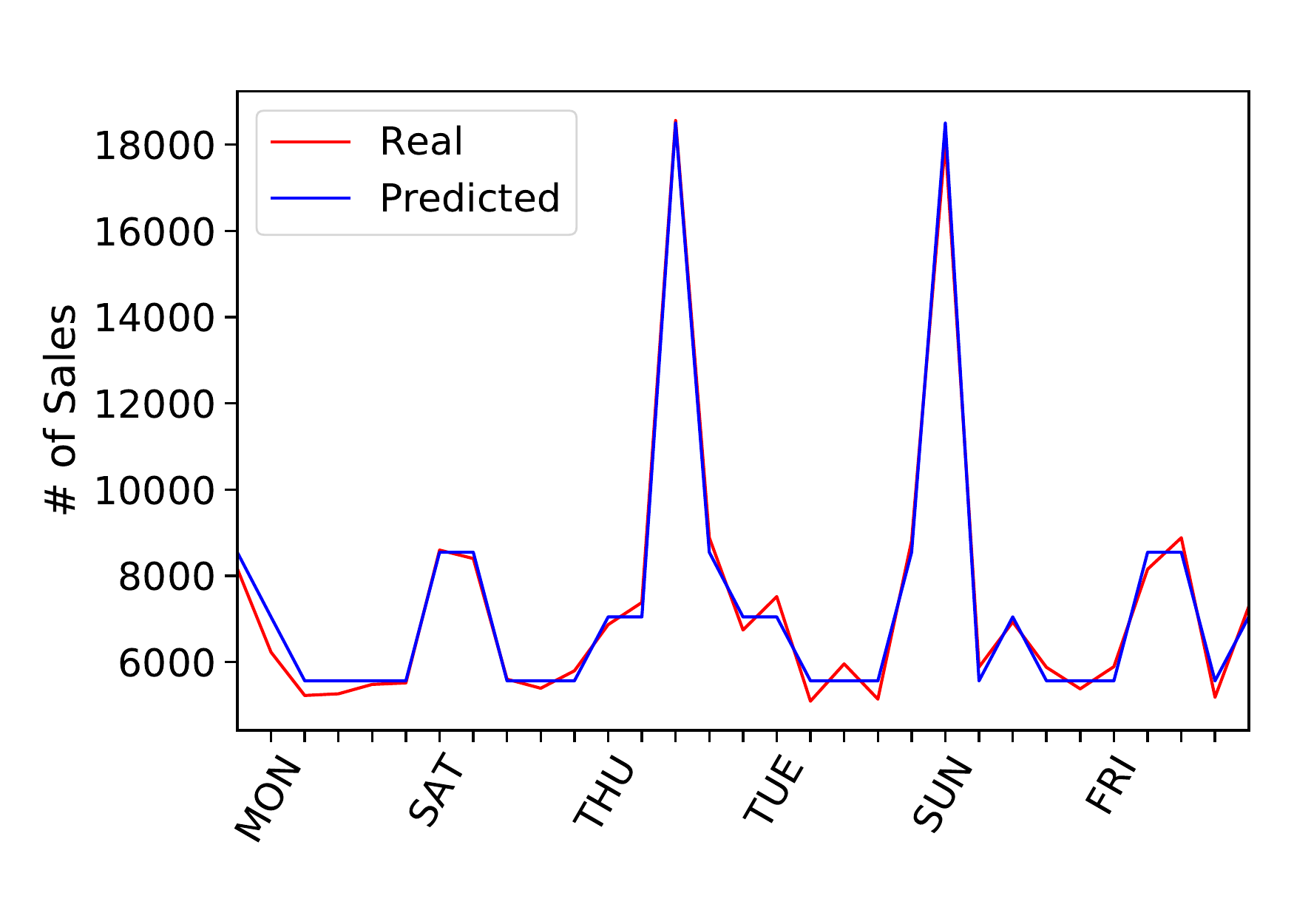}}
    \hfill
    \subfloat[Iteration 5 \label{fig:learnprog5}]{\includegraphics[width=0.5\textwidth]{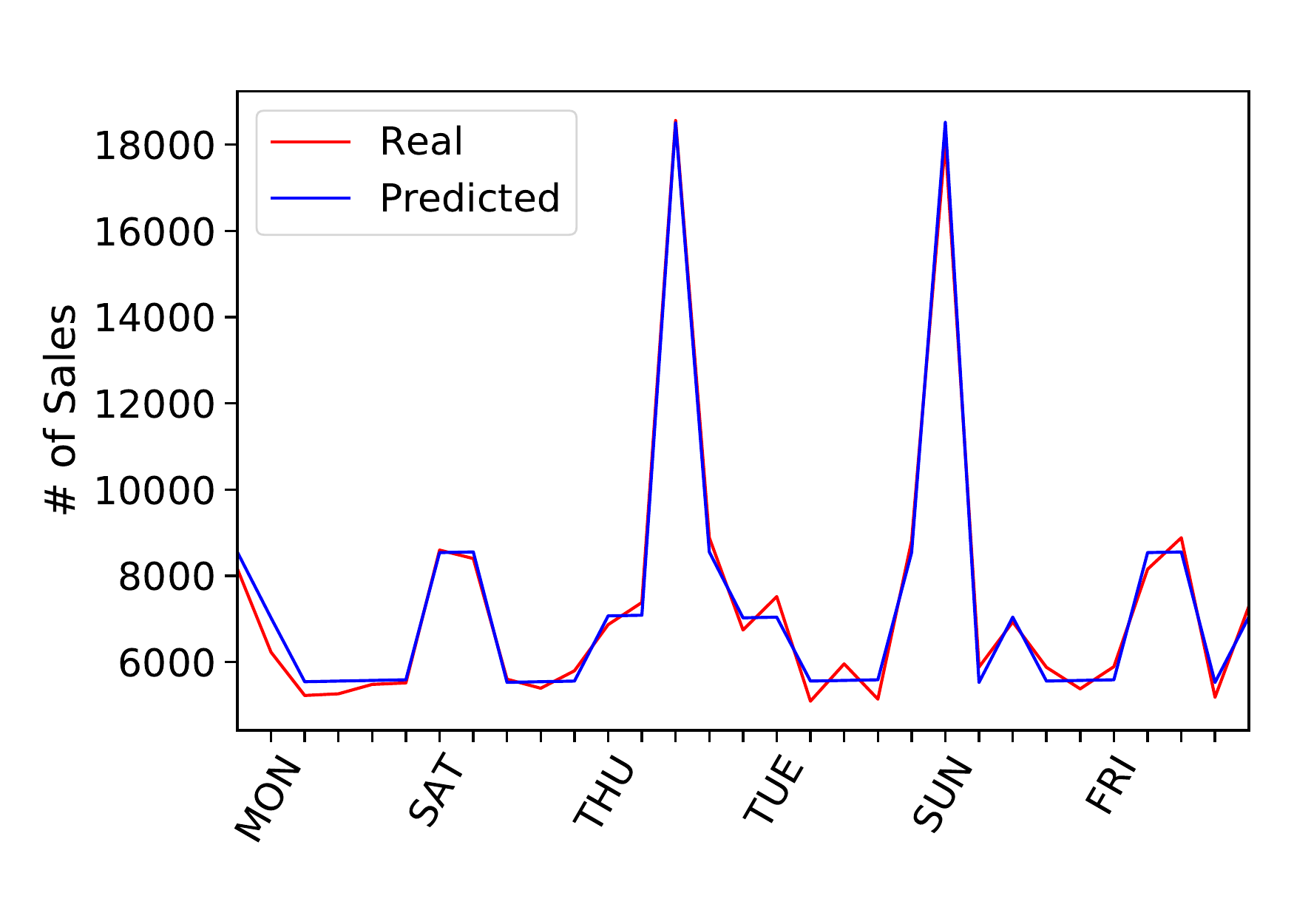}}
    \caption{A sample synthetic time series and predictions of EBLR illustrating the learning process in each iteration.}
    \label{fig:eblr_learning}
\end{figure}

\begin{figure}[!h] 
    \centering
    \includegraphics[scale=0.45]{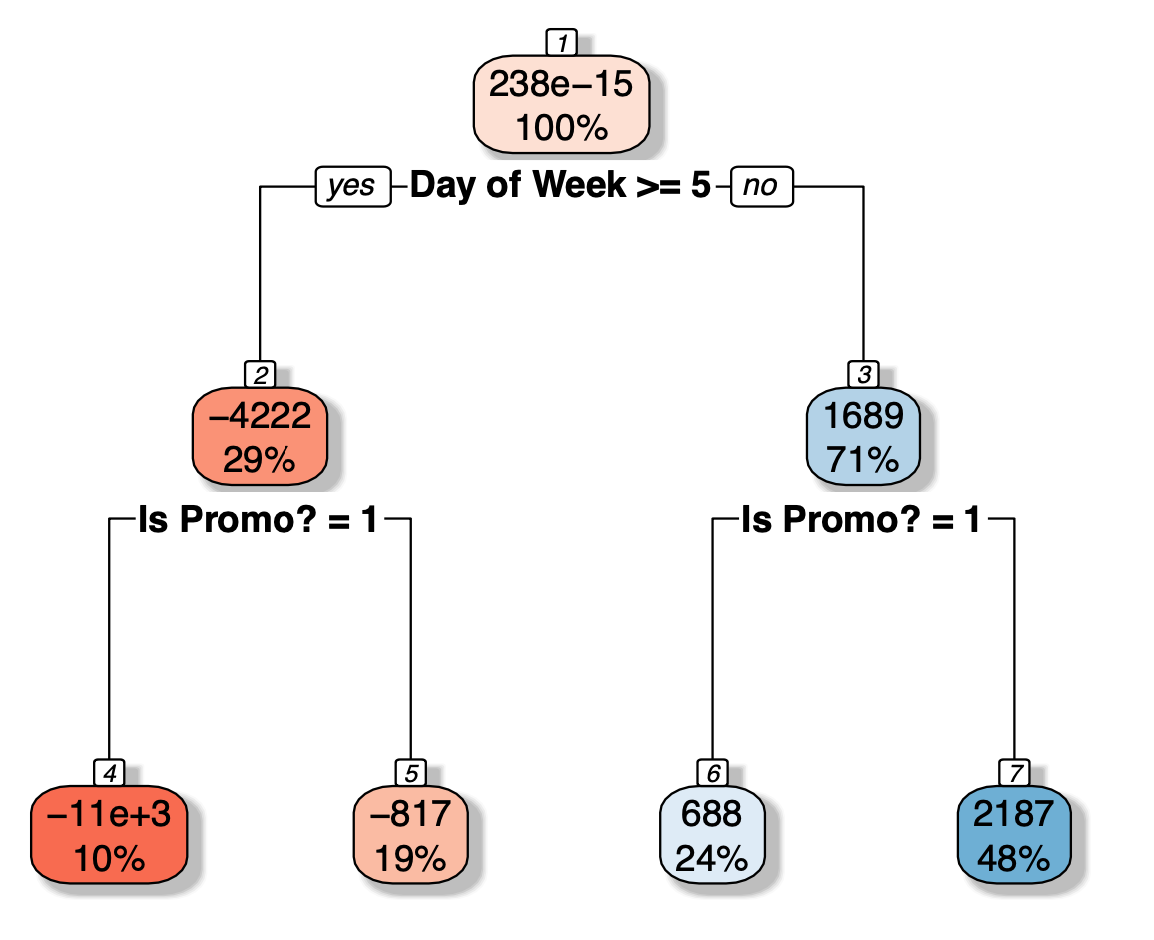}
    \caption{Regression tree trained on the residuals in the first iteration. The top and bottom values in each node respectively illustrates the mean value of the observations and the percentage of the observation falls into the node.}
    \label{fig:first_pruned_tree}
\end{figure}

Figure~\ref{fig:eblr_learning} illustrates the predictions obtained throughout the EBLR iterations. In Figure~\ref{fig:learnprog0}, the red line shows the original time series ($\mathbf{y}$) and the blue line represents our initial guess ($\mathbf{\hat{y}}$) which is the mean of the time series (i.e. we start with an empty feature set). Then the residuals are calculated (the difference between $\mathbf{y}$ and $\mathbf{\hat{y}}$) and a regression tree is fit to the residuals using the data with complete features ($\mathbf{X}$) presented in Figure~\ref{fig:first_pruned_tree}. From the terminal nodes of the tree, the node with the largest mean absolute value is selected (terminal node (4)). Based on this node, a new feature ($\mathbf{\Tilde{X}}$) is created such that it takes value 1 if the observation ends up in the selected terminal node and 0 otherwise. This newly generated feature corresponds to the rule of  {(Is Weekend, Yes) \& (Is Promo, Yes)}. Then, this new feature is added to the feature matrix $\mathbf{X'}$ and the base regressor is retrained. The new predictions ($\mathbf{\hat{y}}$) are illustrated with the blue line in Figure~\ref{fig:learnprog1}. Note that some of the patterns are captured but it still is not enough to discover the underlying model.

% \begin{figure}[!h]
%     \hspace*{-2cm} 
%     \centering
%     \includegraphics[scale=0.6]{images/learning_process.pdf}
%     \caption{EBLR Learning Process}
%     \label{fig:eblr_learning}
% \end{figure}

% Iteration growth (capturing features)

This feature generation and retraining phases are repeated for 5 iterations (each iteration is illustrated in Figure~\ref{fig:eblr_learning}) which provides a good fit to the original time series. Another observation made here is that the early features provide major fixes in the model which are followed by minor fixes as more and more features are added to the model. This also illustrates that the selection of $F^{\max}$ parameter is important and if it is selected too large, EBLR can overfit to the data.

% Here we need a simple example as in the thesis. Two features, they have conditional relation. For example, day of the week and promotion. Promotion only have an effect if it is weekend otherwise it does not.

\subsection{Feature Generation}
Feature generation is an essential part of the EBLR (see Section~\ref{sec:methodology}). The features that are generated from the terminal nodes of the decision tree aim to capture the non-linear relations and interaction effects. Note that these generated features do not have any specific connection to the associated learner and can be used in other prediction models to improve their performance. From this aspect, our method can also be considered as a feature generation method.

Take a dataset of $F$ features. There are a large number of ($2^F - (F+1)$) interaction terms that could be considered. Also, note that each generated feature maps to a set of consecutive decision rules and the variables included in the same set of rules, represents a possible interaction between them. In addition to using these features as is, this kind of intuitive relation also allows us to consider these interaction terms in the prediction models and including them from the beginning for more complex learners.

Here, only the terminal nodes of the decision trees are explored and the node with the largest mean absolute value is selected to generate the features. However, it is possible to apply alternative approaches such as exploring all nodes rather than terminal nodes and selecting many nodes rather than only a single one in each iteration. These approaches might be more prone to over-fitting, however, it speeds-up the feature generation process.

\subsection{Interpretability}
Interpretability is a desired attribute for a prediction method since it brings transparency to the prediction and it increases the reliability of the algorithm. Therefore, prediction algorithms that lack interpretability might not be preferred in some applications or can create hesitancy about making a decision based on the observed results. The biggest advantage of EBLR is its ability to generate non-linear features that are also interpretable. The reason behind its interpretability is that, associated with each generated feature, there is a set of decision rules. These decision rules explicitly shows what this feature represents. Consider the first feature generated in the example provided in Section~\ref{sec:illustration}. This feature corresponds to a decision rule of \{(Is Weekend, Yes) \& (Is Promo, Yes)\} which explicitly points out that promotion has a different effect on the weekends than weekdays. When this feature is added to the model, the corresponding coefficient shows the magnitude and direction of the effect. For example, in the example in Section~\ref{sec:illustration}, it has a positive effect which implies that promotions that are made on weekends are more effective than the weekdays.  

Note that even in this simple example with two features, there are 14 possible interaction terms (7 days $\times$ 2 promo) and EBLR can easily identify that there is no difference among weekdays or weekends, and it is enough to consider whether a day is a weekday or weekend which reduces the total number of interactions to be considered to 4. With an increasing number of features in the data, identification of these interpretable features gets exponentially harder, however, EBLR can easily handle any number of features thanks to the efficient construction and interpretable structure of the decision trees.

\subsection{Extension to Probabilistic Forecasting}
Section~\ref{sec:methodology} describes the proposed algorithm for forecasting a single value (mean). The algorithm can also be extended to probabilistic forecasting. This is done by constructing prediction intervals on the mean estimation based on the empirical error distributions. Assume that we are constructing a prediction interval on the $\hat{y}_i = g(X_i)$, and let $E(\cdot)$ be the distribution function of the errors. Then, the $\alpha\%$ prediction interval constructed as $\hat{y} \pm E^{-1}(\alpha/100)$. This interval can be constructed for any given $\alpha$ based on the required confidence. %One disadvantage of this approach is that for each value of $\hat{y}_i$, the half-width of the intervals become the same. This implicitly implies that there is an equal uncertainty over all parts of the feature matrix. To alleviate this issue, we pursue another approach. Specifically, we apply our method until all the features are generated, and, once we have the complete feature set, instead of only fitting a linear regression (or LASSO regression), we also fit quantile regressions (or penalized quantile regressions) for various quantiles. Then, for a given feature matrix $X_i$, we can predict associated quantiles based on each model.

% \begin{itemize}
%     \item Deep Learning example + bad performance with small data
% \end{itemize}
%%%%%%%%%%%%%%%%%%%%%%%%%%%%%%%%%%%%%
\section{Numerical Experiments}\label{sec:num_experiments}
This section provides a set of experiments to demonstrate the prediction performance of EBLR in both deterministic and stochastic settings. Before presenting the results, the experimental setup is provided. This includes the information regarding datasets, performance metrics, model settings and implementation details.

\subsection{Experimental Setup}
EBLR is created using both \texttt{scikit-learn} and \texttt{r-forecast} packages from two programming languages and is provided in \cite{irogi2020eblr}.
%The benefit of creating EBLR in this format was that \texttt{r-forecast} contains a robust decision tree algorithm, with cross validation post pruning built-in. This does not exists inside \texttt{scikit-learn}. 
The experiments are conducted on a 2.7 GHz dual-core i5 processor with 8GB of RAM, using the Python programming language. 

%However, the remainder of the Python environment was well suited to create EBLR, run experiments, and record results in an efficient manner.

\subsubsection{Datasets}
In this work, three datasets are utilized in the experiments: synthetic, Rossman \cite{data_rossmann_old} and Turkish electricity\footnote{https://seffaflik.epias.com.tr/transparency/tuketim/gerceklesen-tuketim/gercek-zamanli-tuketim.xhtml}. Summary information regarding these datasets can be found in Table \ref{tab:dataset_statistics}.

\setlength{\tabcolsep}{4pt} % adjust column separation in table
\renewcommand{\arraystretch}{1.3} % adjust row separation in table
\begin{table}[!h]
    \centering
    \caption{Descriptive information and statistics on the datasets.}
    \label{tab:dataset_statistics}
    \scalebox{0.9}{
    \begin{tabular}{lccc}
    \toprule
 & \textbf{Synthetic} & \textbf{Rossmann} & \textbf{Turkish Electricity} \\ 
 \midrule
\# features                      & 1                  & 6                 & 2                            \\
\# time covariates               & 1                  & 53                & 31                           \\
\# time series                   & 1                  & 100               & 1                            \\
\# avg data points / time series & 2048               & 918.1             & 8760                         \\
Granularity                      & Daily              & Daily             & Hourly                       \\
Seasonality                      & None               & None              & Daily                        \\
Forecast Horizon                 & 14 days            & 28 days           & 14 hours                     \\
\# of tests                      & 25                 & 100               & 25                           \\ 
\bottomrule
\end{tabular}
}
\end{table}

\begin{figure}[!htp]
    \centering
    \includegraphics[width=0.5\textwidth]{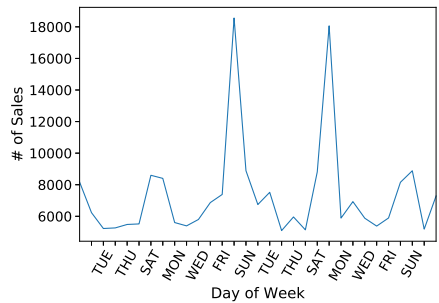}
    \caption{A one-month training snapshot of the synthetic dataset, showcasing linear features (small peaks) and two non-linear interactions (large peaks).}
    \label{fig:sample-data}
\end{figure}

\noindent\textbf{Synthetic dataset.} The first dataset is a synthetically generated dataset that mimics a simple version of daily sales of a retail store. Figure~\ref{fig:sample-data} demonstrates a sample from this synthetic dataset. Note that the same approach is used in Section~\ref{sec:illustration} for illustration purposes.

\noindent\textbf{Rossman dataset.}
The Rossmann sales data is a feature-rich dataset, containing features related to the date, number of customers, holiday-specific information, promotional information, and store closures. The sales of an individual store is set to the target variable, and the customer covariate was removed because it is related to the target variable, and requires to be forecasted. The date feature is further decomposed into the day of the week, day of the month, month of the year, and year features for enriching the feature space.

\noindent\textbf{Turkish electricity dataset.} Finally, the highly seasonal Turkish electricity consumption dataset is used. Here, the prediction focus is on the electricity consumption of Turkey's most populated city's (Istanbul). This target variable is created by interpolating the total electricity consumption in Turkey to Istanbul (based on monthly factors). This dataset is decomposed into hourly consumption rates, with daily seasonality. Then, the date is decomposed into the day of the week, month and year variables. Moreover, the data contains daily high and low temperatures of Istanbul which is an important determinant of electricity consumption.

\subsubsection{Performance Metrics}
In this work, three performance metrics are considered similar to \citep{salinas2020deepar}. Point forecasting performance is compared in terms of the normalized root mean squared error (NRMSE) and the normalized deviation (ND). Explicit expressions for the metrics are provided as follows:
% def nrmse(y_pred, y_real):
%   return np.sqrt(np.mean((y_pred - y_real)**2)) / np.mean(y_real)
% def nd(y_pred, y_real):
%   return np.sum(np.abs(y_pred - y_real)) / np.sum(y_real),
\begin{align}
    \text{NRMSE}(y, \hat{y}) &= \frac{\sqrt{\frac{1}{N} \sum_{i=1}^{N}(\hat{y_i} - y_i)^2}}{\frac{1}{N} \sum_{i=1}^{N} \vert y_i \vert} \\
    \text{ND}(y, \hat{y}) &= \frac{\sum_{i=1}^{N}\vert\hat{y_i} - y_i\vert}{\sum_{i=1}^{N} \vert y_i \vert}
\end{align}

For the probabilistic forecasting, the models are evaluated on the weighted scaled pinball loss at the 0.05, 0.25, 0.50, 0.75 and 0.95 quantiles. To combine these five metrics into one general metric, the mean of the quantile losses is also reported. For any quantile $\rho$, the weighted scaled pinball loss is calculated as follows:
\begin{align}
    \text{WSPL}_\rho(y, \hat{y}^\rho) &= \frac{\sum_{i=1}^{N} \max\{\rho(y_i - \hat{y_i}^\rho), (1-\rho)(\hat{y_i}^\rho - y_i)\}}{\sum_{i=1}^{N} \vert y_i \vert}
\end{align}
% def quantile_loss(quant, y_real, y_pred):
%     err = y_real - y_pred
%     return np.mean(np.maximum(quant * err, (quant - 1) * err)) / np.mean(y_real)

\subsubsection{Model Settings}
In the point forecasting experiments, EBLR is compared against two naive methods: linear regression (LR) and a baseline method (Mean), statistical models: ARIMAX, and ensemble methods: gradient boosting regressor (GBR) \cite{friedman2001greedy}, random forest regressor (RF) \cite{breiman2001random}. LASSO penalty is selected based on 5-fold cross-validation. For ARIMA based models (e.g. ARIMAX), a step-wise algorithm is used to tune the model (see ~\citep{hyndman2007automatic} for details). In the case of the seasonal Turkish electricity dataset, the seasonal parameters are included in the ARIMAX model as well. This is because the Turkish electricity dataset is inherently seasonal~\citep{ilic2020augmented}. For GBR and RF, a single parameter setting is utilized without performing hyperparameter tuning, and this setting is used consistently across all datasets. The parameters for GBR and RF can be found in Table~\ref{tab:baseline_hyperparameters}.
\begin{table}[!h]
    \centering
    \caption{The hyperparameter setting used in the experiments for gradient boosting regression (GBR) and random forest (RF).}
    \label{tab:baseline_hyperparameters}
    \scalebox{0.9}{
    \begin{tabular}{lccc}
    \toprule
    & \textbf{GBR} & \textbf{RF} \\ 
    \midrule
    \# of trees & 100 & 100 \\
    Splitting Criterion & Friedman MSE & MSE \\
    Max Depth & 3 & $\infty$ \\ 
    Loss Function & Least Squares Regression & N/A \\ 
    Learning Rate & 0.1 & N/A \\ 
    \bottomrule
    \end{tabular}
    }
\end{table}

Moreover, for EBLR, the parameter setting for each dataset are specified in Table~\ref{tab:eblr_hyperparameters}. Only a small number of parameters were selected in the synthetic data set since it was known that there were only a few underlying features. Then, 50 features were arbitrarily chosen to be learned in the Rossmann data set. When this setting was reused on the Turkish dataset, it was found that the model was still learning, therefore the number of features was doubled. All the initial complexity parameters where chosen such that they were significantly small.

\begin{table}[!h]
    \centering
    \caption{EBLR parameters used in the experiments for three datasets.}
    \label{tab:eblr_hyperparameters}
    \scalebox{0.9}{
    \begin{tabular}{lccc}
    \toprule
    & \textbf{Synthetic} & \textbf{Rossmann} & \textbf{Turkish Electricity} \\ 
    \midrule
    \# of features & 5 & 50 & 100 \\
    Initial Complexity Parameter & 0.001 & 0.001 & 0.001 \\ 
    \bottomrule
    \end{tabular}
    }
\end{table}

In all the probabilistic forecasting experiments, EBLR was compared against both ensemble methods as found in point forecasting. In order to generate probabilistic results, the loss function was adjusted in the gradient boosted regressor model to use quantile loss, and a model was trained for each quantile. In addition, in random forest, the quantiles were collected based on the individual regressors. The other models use a more naive forecasting approach, by training prediction intervals based on the training residuals. All the experiments were repeated with the same feature setups, however, instead the 5\%, 25\%, 50\%, 75\% and 95\% quantile predictions were collected. As expected, it is seen that all of the 50\% quantiles perform similarly to the ND scores.

\subsection{Improvement over Base Regressors}
This section presents an experiment on EBLR's contribution on improving the prediction performances of linear regression, LASSO regression and ARIMA methods. Table~\ref{tab:base_regression} illustrates the average forecasting performances of these methods with and without EBLR on synthetic dataset and Rossmann dataset. From these results, it can observed that utilizing EBLR provides significant decrease in the NRMSE and ND for both linear regression, LASSO and ARIMA methods, especially for the synthetic dataset. Note that since these base regressors are linear, they fail to capture the interaction effect in the synthetic data whereas incorporating these base regressors in EBLR allows the generation of interaction features which reduces the error significantly. Since EBLR provides similar results with all base regressors, we utilize LASSO regression as the base regression in the future experiments with the aim of preventing potential over-fitting.

\begin{table}[!h]
\caption{Point forecasting results for linear regression, LASSO regression and ARIMA methods with their EBLR version.}
\label{tab:base_regression}
\scalebox{0.85}{
\begin{tabular}{llcccccc}
\toprule
& & \multicolumn{2}{c}{Linear Regression} & \multicolumn{2}{c}{LASSO} & \multicolumn{2}{c}{ARIMA} \\
& & w/o EBLR        & w EBLR       & w/o EBLR  & w EBLR & w/o EBLR  & w EBLR \\
\midrule
Synthetic & NRMSE &  0.3265                   & 0.0471         & 0.3264      & 0.0472  & 0.3383       & 0.0471  \\
& ND    & 0.2474                    & 0.0383         & 0.2473      & 0.0384  & 0.2626       & 0.0383  \\
\midrule
Rossmann & NRMSE & 0.1731 & 0.1528 & 0.1750      & 0.1543 & 0.1754 & 0.1613  \\
& ND & 0.1252 & 0.1085 & 0.1270 & 0.1088 & 0.1284  & 0.1164  \\ 
\bottomrule
\end{tabular}
}
\end{table}

\subsection{Point Forecasting}\label{sec:point_res}
\begin{figure}[!h]
    \centering
    \subfloat[Synthetic dataset \label{fig:synthetic_point}]{\includegraphics[width=0.5\textwidth]{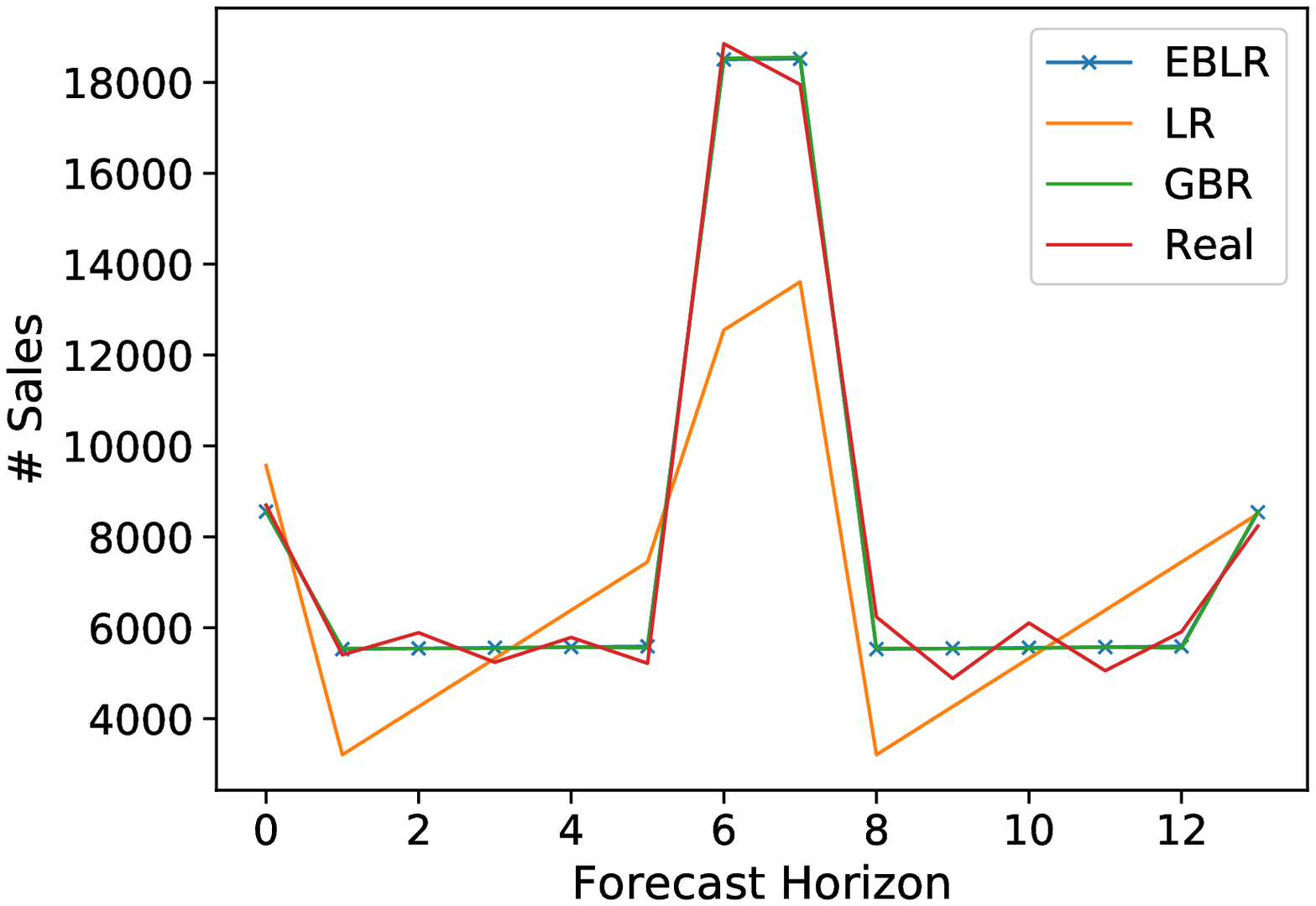}}
    \hfill
    \subfloat[Rossmann \label{fig:rossman_point}]{\includegraphics[width=0.5\textwidth]{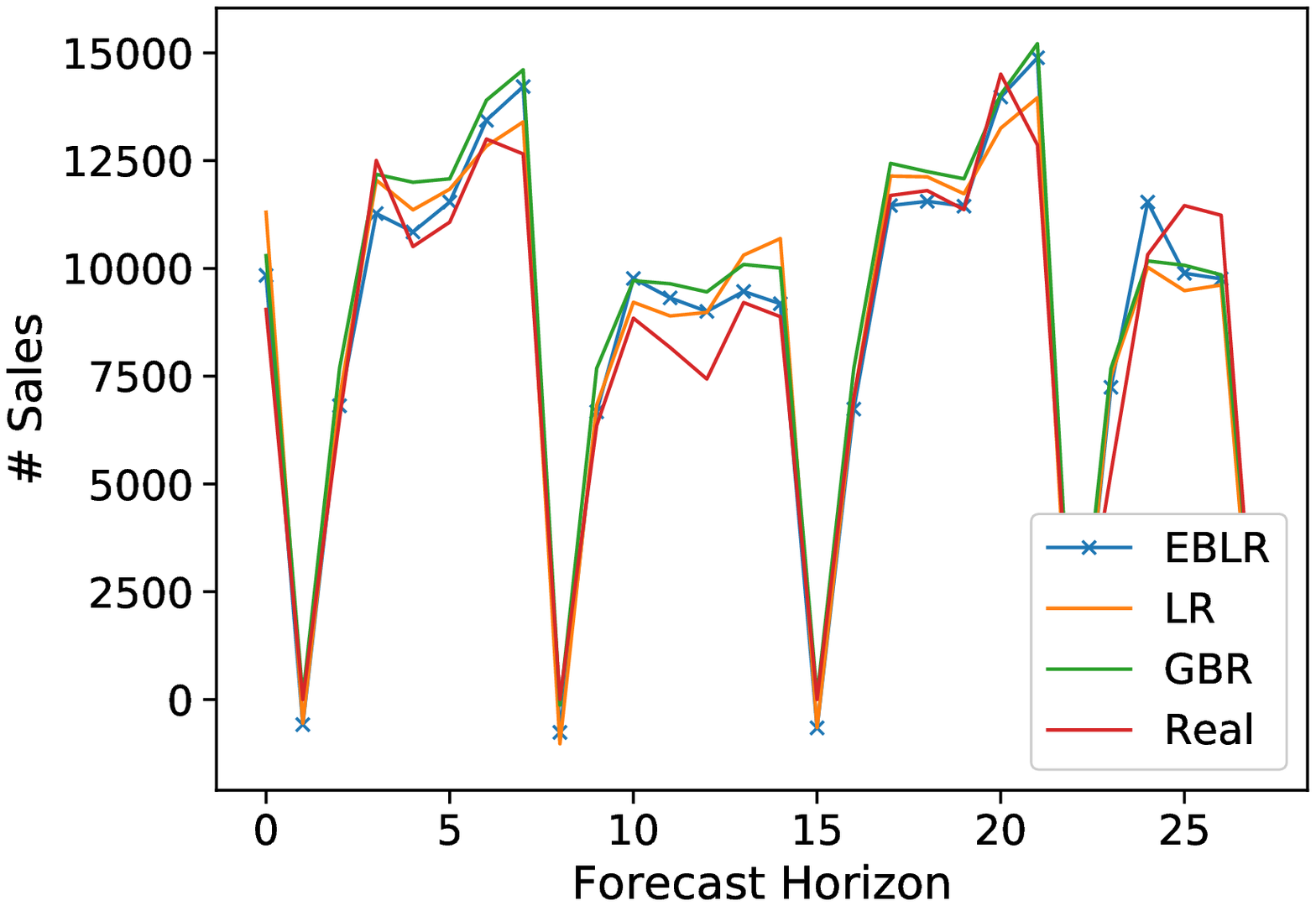}}\\
    \subfloat[Turkish electricity \label{fig:turkish_point}]{\includegraphics[width=0.5\textwidth]{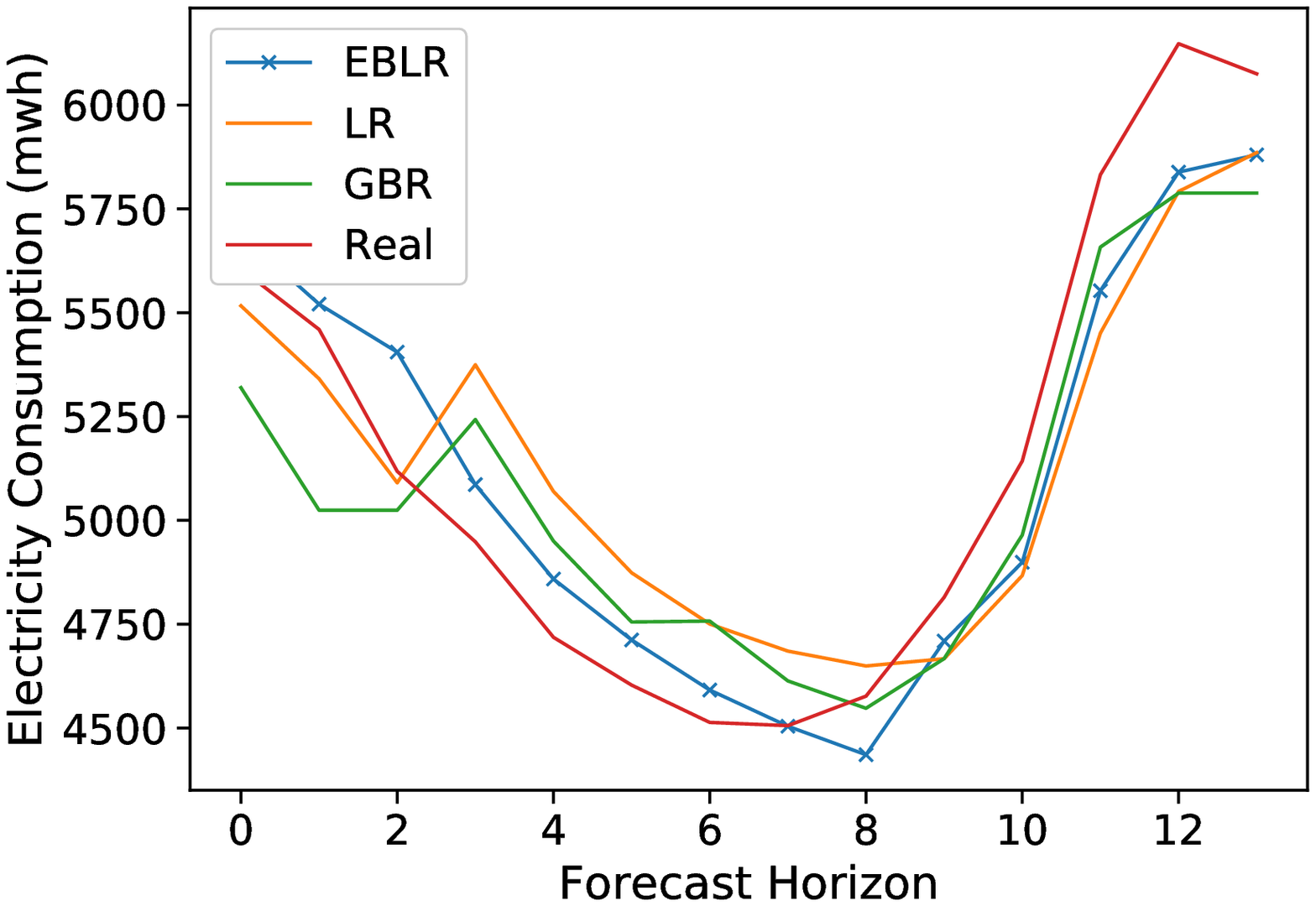}}
    \caption{Three different point forecasting tests of EBLR, linear regression (LR), gradient boosting regressor (GBR). The order (from left to right, top to bottom) include a 14 period forecast with respective predictions, a 28 period sales forecast, and a 14 hour electricity consumption forecast.}
    \label{fig:pointForecasts}
\end{figure}

% The first set of experiments are run on a synthetic data set, designed with non-linear features. Here, We ran an augmented out-of-sample test with the aim of capturing the next 14 forecast predictions from the end of the training data set \citep{ilic2020augmented}. In order to ensure robustness of the results, the experiment is repeated 25 times. 
The prediction performances over three datasets are provided in Table~\ref{tab:point_forecasting}, moreover, Figure~\ref{fig:pointForecasts} illustrates the predictions of four important regressors on single time series from each of these three datasets. The first set of experiments are run on a synthetic data set and the results are presented in the second and third columns of Table~\ref{tab:point_forecasting}. From these results, three specific observations can be made: (1) EBLR significantly improves the performance of linear regression (decrease NRMSE from 0.3265 to 0.0472) by introducing 5 additional features that captures the important interactions. (2) EBLR outperforms all primitive regression algorithms including ARIMAX, linear regression and naive baseline. (3) EBLR provides comparable performances to the ensemble methods.

% All ensemble models were able to capture the added features and account for the uncertainty. In Section \ref{subsec:feat_imp}, we look deeper into EBLRs generated features. Linear regression and ARIMAX were unable to capture this relationship due to the non-linear feature, and our baseline mean model performed the worst. 

% Plot 1 forecast / dataset (EBLR+GBR+LR or all if doable)

% We see that EBLR performed exactly as intended for the synthetic data set. With this realization, we have extended our experiments to two additional data sets. 
Secondly, we focus on the Rossmann dataset where the results are presented in the fourth and fifth columns of Table~\ref{tab:point_forecasting}. In the Rossmann dataset, EBLR performs slightly worse than GBR, but once again outperforms ARIMAX, linear regression and naive baseline models. When EBLR is compared to random forest, a more interesting relationship is revealed. EBLR has a lower NRMSE whereas random forest has a lower ND. This means that EBLRs predictions tend to be more consistent, whereas random forest on average makes a more accurate prediction, however the poor predictions are much worse than EBLRs. Finally, the last two columns of Table~\ref{tab:point_forecasting} demonstrates the results on Turkish electricity dataset, where, consistent with the previous datasets, EBLR outperforms ARIMAX, linear regression, and naive baseline models. In addition, EBLR also outperforms GBR, but not RF. 

Overall, EBLR provides substantial improvement to the linear regression by incorporating additional features. Moreover, even though EBLR is completely interpretable, it provides comparative performance to the state-of-the-art ensemble-based methods.  
% Store tables inside a google sheet
\begin{table}[!h]
    \centering
    \caption{Point forecasting results over all three datasets for EBLR, random forest (RF), gradient boosting regression (GBR), ARIMAX, linear regression and naive baseline methods.}
    \label{tab:point_forecasting}
    \scalebox{0.9}{
    \begin{tabular}{ccccccc}
    \hline
    \multirow{2}{*}{Model} & \multicolumn{2}{c}{\textbf{Synthetic}} & \multicolumn{2}{c}{\textbf{Rossmann}} & \multicolumn{2}{c}{\textbf{Turkish Electricity}} \\ \cline{2-7} 
& NRMSE              & ND                & NRMSE             & ND                & NRMSE                   & ND                     \\ \hline
EBLR                   & 0.0472             & 0.0384            & 0.1528            & 0.1085            & 0.0428                  & 0.0348                  \\
RF                     & 0.0471             & 0.0384            & 0.1670             & 0.1017            & 0.0297                  & 0.0234                 \\
GBR                    & 0.0471             & 0.0384            & 0.1507            & 0.0986            & 0.0581                  & 0.0466                 \\
ARIMAX                 & 0.3384             & 0.2626            & 0.1754            & 0.1284            & 0.0500                  & 0.0297                 \\
Linear Regression      & 0.3265             & 0.2474            & 0.1750             & 0.1270             & 0.0530                   & 0.0424                 \\
Baseline               & 0.4778             & 0.3007            & 0.5880             & 0.4155            & 0.1171                  & 0.1019                 \\ \hline
\end{tabular}
    }
\end{table}

% What we did to further these results is we have extracted the features generated inside EBLR, and placed them as features on SARIMAX. When we did this, we were able to increase SARIMAX's performance by over ##% on average. This shows that EBLR expands beyond a simple ensemble method. EBLR is actually able to generate features to improve other models.

\subsection{Probabilistic Forecasting}\label{sec:prob_res}
\begin{figure}[!h]
    \centering
    \subfloat[EBLR \label{fig:synth_prob_ebr}]{\includegraphics[width=0.5\textwidth]{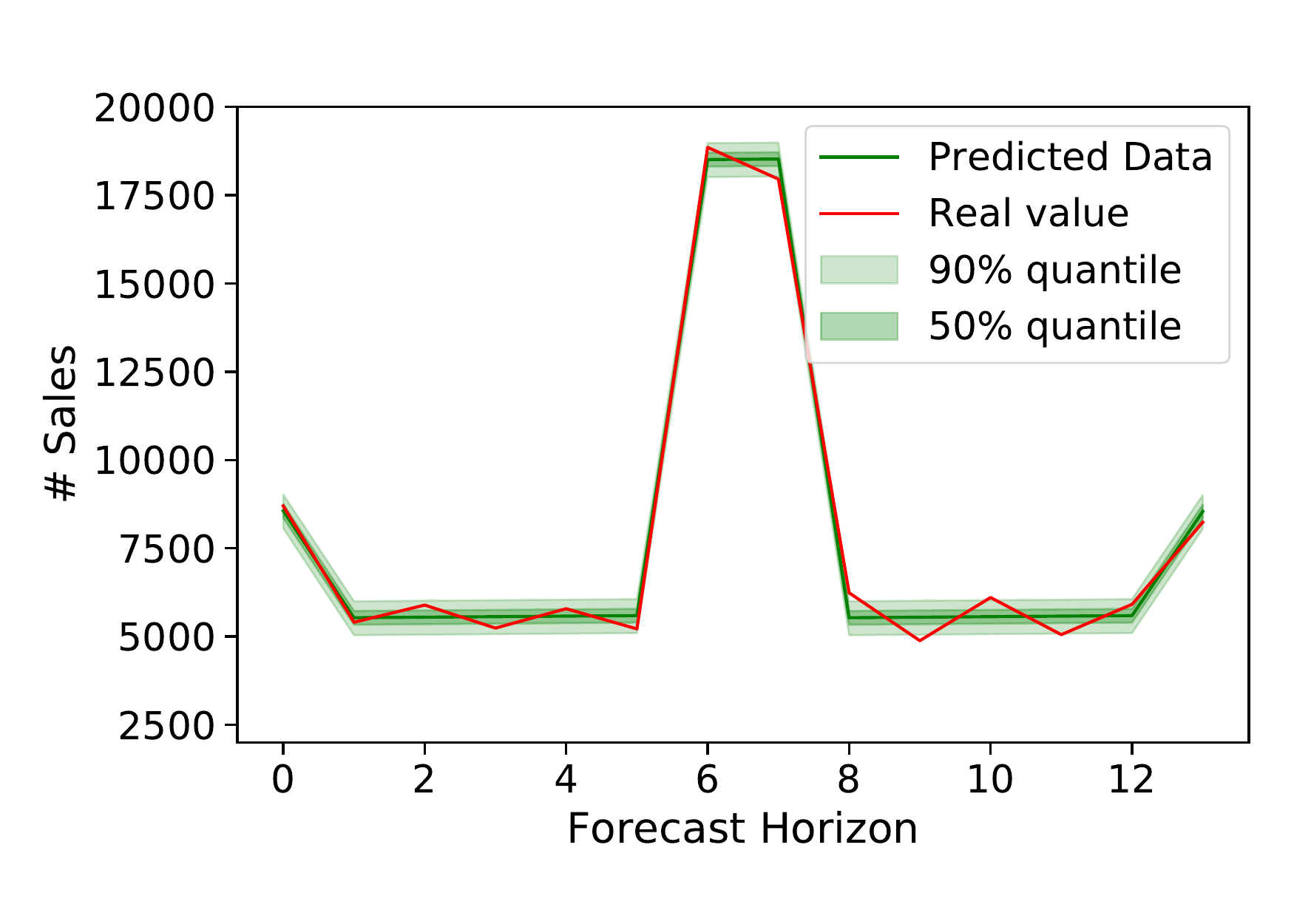}}
    \hfill
    \subfloat[GBR \label{fig:synth_prob_gbr}]{\includegraphics[width=0.5\textwidth]{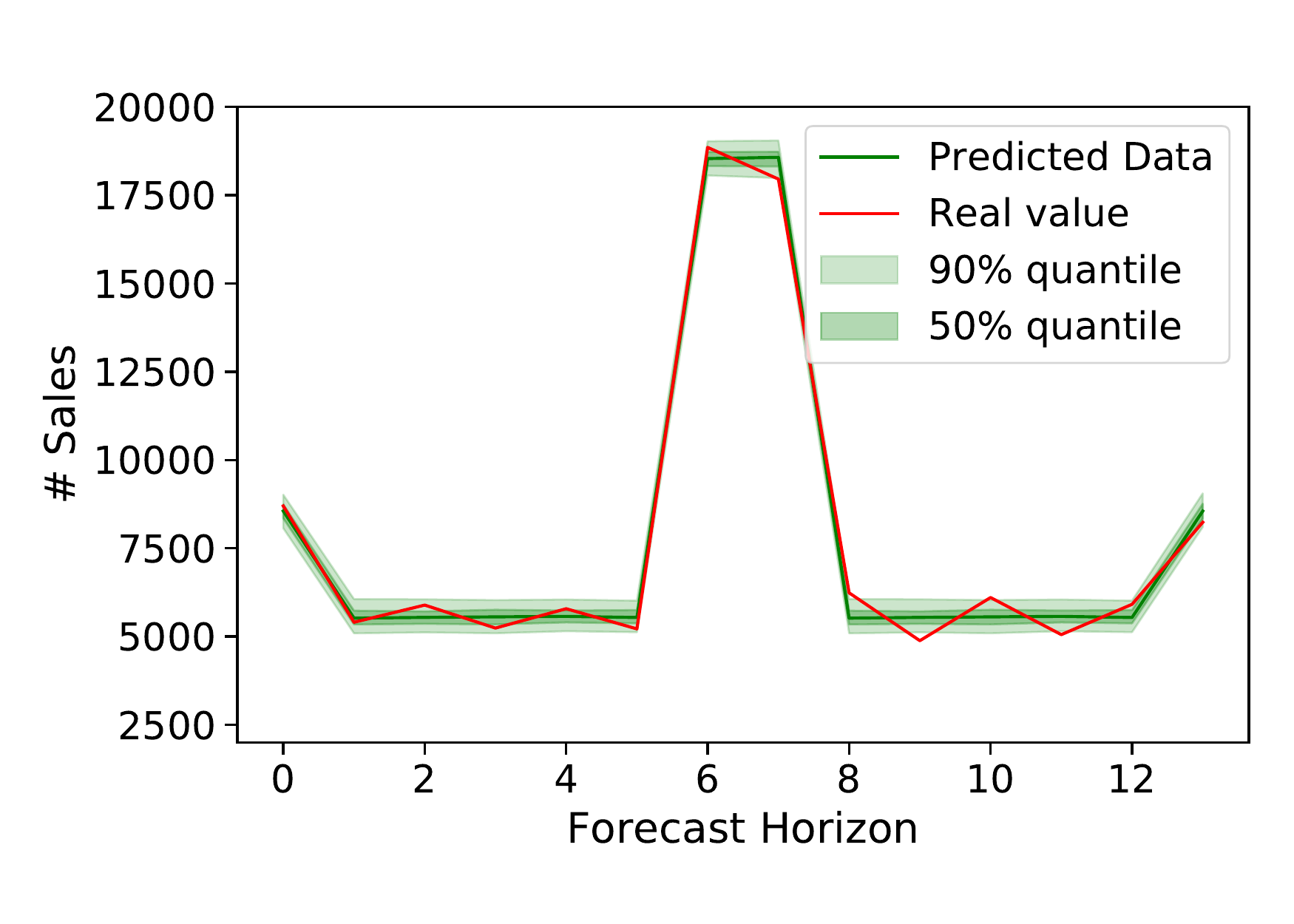}}\\
    \subfloat[ARIMAX \label{fig:synth_prob_arima}]{\includegraphics[width=0.5\textwidth]{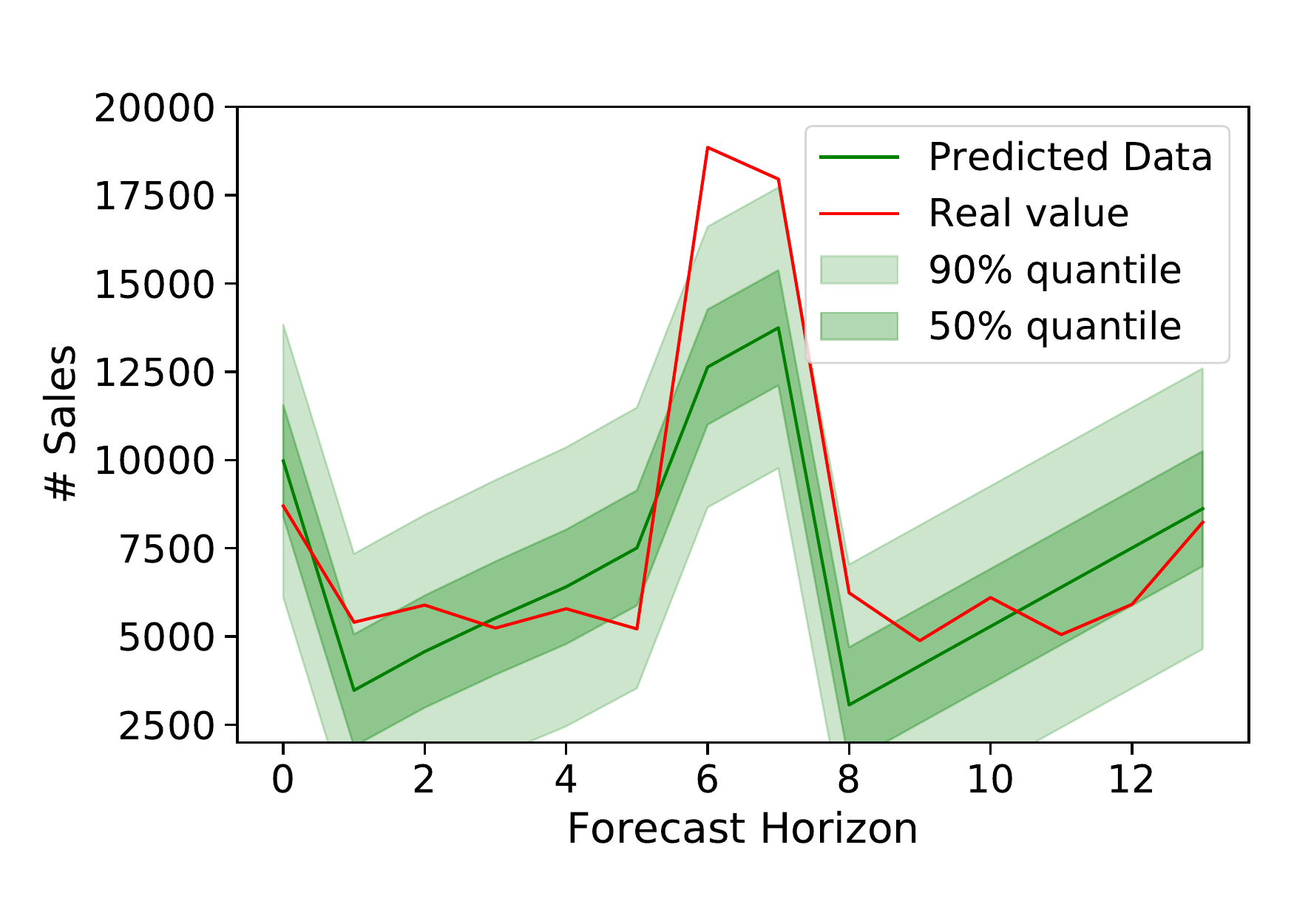}}
    \caption{Three different probabilistic forecasting interval results on a sample forecasting horizon. Each forecast horizon contains the median forecast (dark green line), the 50\% forecast interval (medium green fill), and 90\% forecast interval (light green fill).}
    \label{fig:probForecastsSynth}
\end{figure}

This section presents probabilistic forecasting results on the three datasets. The first set of results focus on the synthetic dataset for which the summary statistics are presented in Table \ref{tab:synth_res_prob}, and example probabilistic forecasts for EBLR, GBR and ARIMAX models are provided in Figure~\ref{fig:probForecastsSynth}. Overall, EBLR performs the same as GBR and outperforms all other competitors. EBLR captures the tails very well as well as the median predictions. This suggests that EBLR extends to probabilistic forecasting successfully.

%This is to be expected, as EBLR was able to determine the probability distribution of the residuals, $\mathcal{N}(\mu, \sigma)$, while capturing the features. 

%similarly to the other two ensemble methods, and much better than linear regression, ARIMAX and naive baseline methods.  

% We notice the other ensemble methods are slightly more accurate, but this can be attributed to the simple method of generating the intervals.In order to generate a more generalized intervals, we can look into more intricate ways to generate the prediction intervals. This has been left as an area for future research.

\begin{table}[!h]
\centering
\caption{Synthetic probabilistic forecasting results for the five specified quantiles and their combined mean for EBLR, random forest (RF), gradient boosting regression (GBR), ARIMAX, linear regression and naive baseline methods.}
\label{tab:synth_res_prob}
\begin{tabular}{ccccccc}
\hline
\multirow{2}{*}{Model} & \multicolumn{6}{c}{\textbf{WSPL($\rho$)}}                \\ \cline{2-7} 
        & 0.05   & 0.25   & 0.50   & 0.75   & 0.95   & Mean   \\ \hline
EBLR    & 0.0051 & 0.0157 & 0.0192 & 0.0153 & 0.0048 & 0.0120 \\
GBR     & 0.0053 & 0.0156 & 0.0192 & 0.0152 & 0.0048 & 0.0120 \\
RF      & 0.0168 & 0.0186 & 0.0192 & 0.0187 & 0.0169 & 0.0180 \\
MEAN    & 0.0397 & 0.0848 & 0.1504 & 0.1829 & 0.0976 & 0.1111 \\
ARIMAX  & 0.0337 & 0.0986 & 0.1246 & 0.1175 & 0.0473 & 0.0843 \\ \hline\end{tabular}
\end{table}

% Moving to the other data sets, we find that EBLR is better than simple linear models, and reaches the tier of GBR and RF. Each quantile has a separate GBR model trained to minimize the respective quantile loss. In order to perform the performance of EBLR, a separate model could be trained for each quantile instead of bootstrapping the residuals.

\begin{figure}[!h]
    \centering
    \subfloat[EBLR \label{fig:rossman_prob_eblr}]{\includegraphics[width=0.5\textwidth]{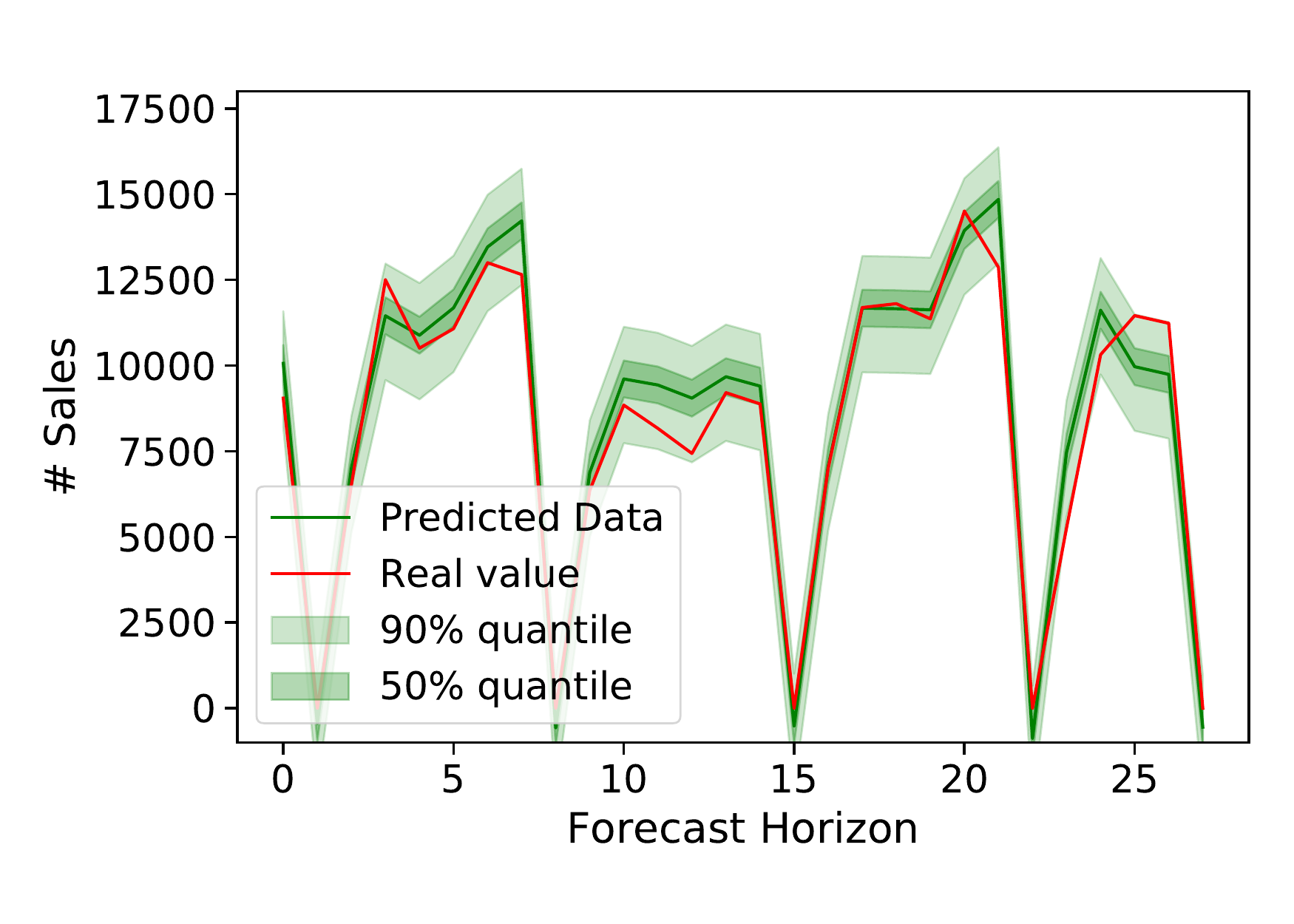}}
    \hfill
    \subfloat[GBR \label{fig:rossman_prob_gbr}]{\includegraphics[width=0.5\textwidth]{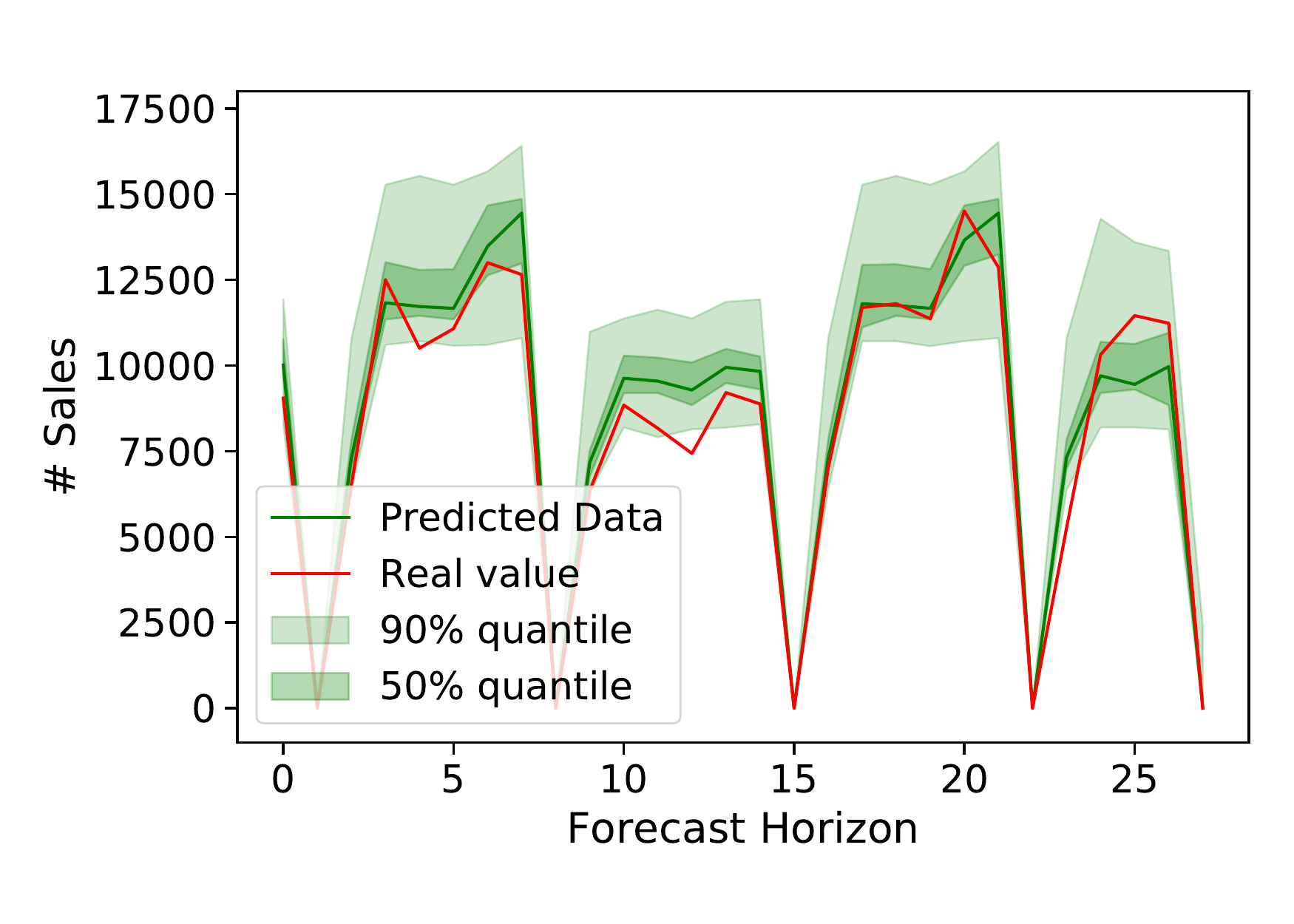}}\\
    \subfloat[ARIMAX \label{fig:rossman_prob_arima}]{\includegraphics[width=0.5\textwidth]{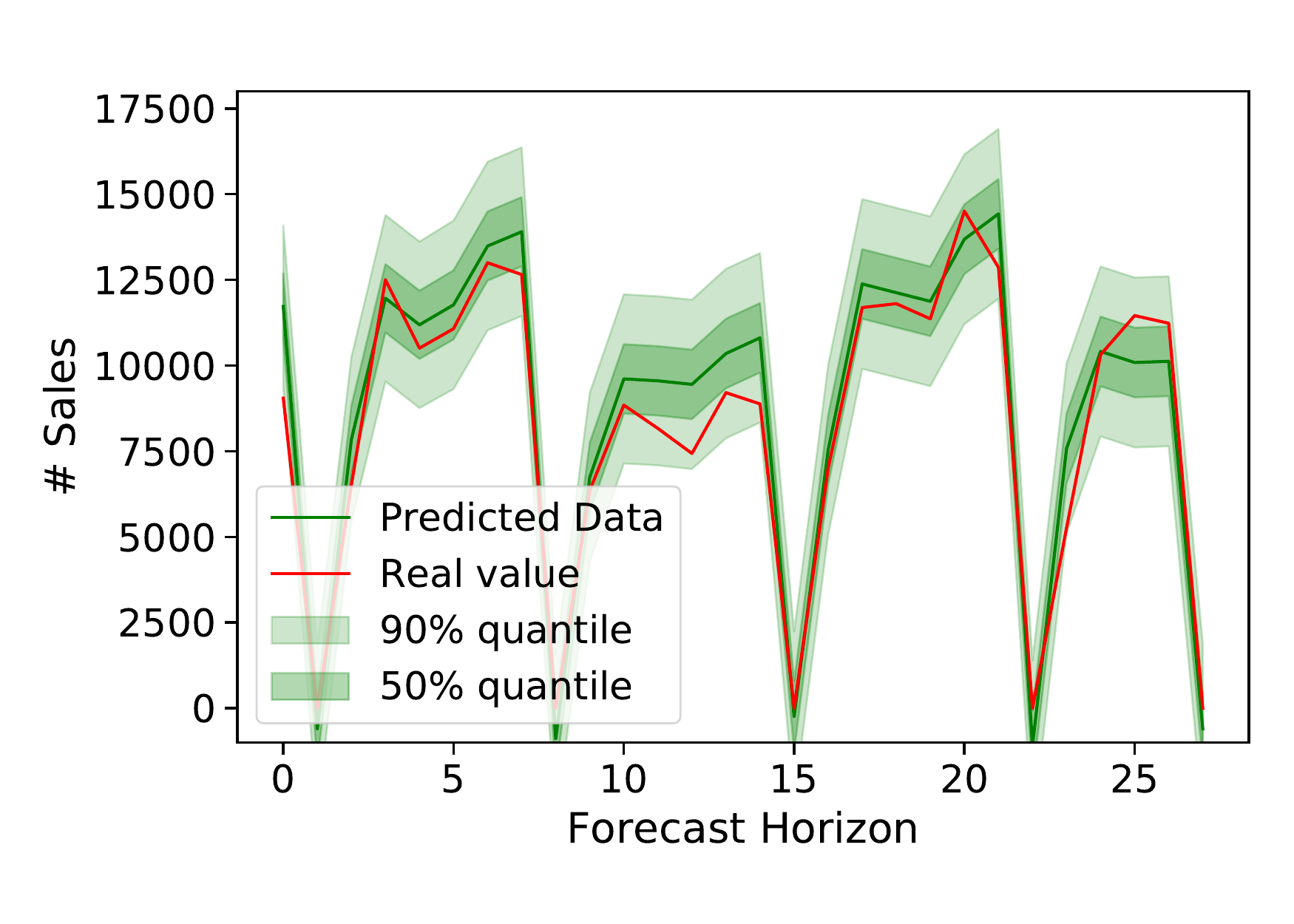}}
    \caption{Three different probabilistic forecasting interval results on a 28 day forecast horizon for a selected Rossmann store. Each forecast horizon contains the median forecast (dark green line), the 50\% forecast interval (medium green fill), and 90\% forecast interval (light green fill).}
    \label{fig:probForecastsRoss}
\end{figure}

Secondly, the performances of the subjected methods are evaluated on the Rossmann dataset. Table~\ref{tab:rossman_prob_res} illustrates the performances of the considered methods and Figure \ref{fig:probForecastsRoss} provides an illustration of the predictions for EBLR, GBR and ARIMAX methods. Here, EBLR performs worse than GBR, however, it outperforms all other methods. One important observation is that the relative performance of EBLR to GBR is better in extreme quantiles. This indicates that EBLR accounts for unlikely outcomes more than GBR.

\begin{table}[!h]
\centering
\caption{Three different probabilistic forecasting interval on a sample Rossmann store 28 day forecast. Each forecast horizon contains the median forecast (dark green line), the 50\% forecast interval (medium green fill), and 90\% forecast interval (light green fill).}
\label{tab:rossman_prob_res}
\begin{tabular}{ccccccc}
\hline
\multirow{2}{*}{Model} & \multicolumn{6}{c}{\textbf{WSPL($\rho$)}}                 \\ \cline{2-7} 
       & 0.05   & 0.25   & 0.50   & 0.75   & 0.95   & average \\ \hline
EBLR   & 0.0185 & 0.0466 & 0.0539 & 0.0409 & 0.0140 & 0.0348 \\
GBR    & 0.0161 & 0.0401 & 0.0474 & 0.0376 & 0.0162 & 0.0315 \\
RF     & 0.0226 & 0.0421 & 0.0527 & 0.0429 & 0.0144 & 0.0349 \\
MEAN   & 0.0626 & 0.2032 & 0.2078 & 0.1575 & 0.0556 & 0.1373 \\
ARIMAX  & 0.0195 & 0.0538 & 0.0627 & 0.0506 & 0.0182 & 0.0410 \\ \hline
\end{tabular}
\end{table}

Lastly, Table~\ref{tab:turkish_prob_res} and Figure \ref{fig:probForecastsTurk} respectively illustrate the experiment results and example predictions of different methods for the Turkish electricity dataset. From these results, in contrast to the Rossman dataset, EBLR outperforms GBR at all quantiles. This could be attributed to GBR overfitting the data set and starting to include noise from training data into the model. EBLR is able to outperform the baseline model as well as outperforming the ARIMAX model. A unique observation is that, since the ARIMAX model is given extra seasonal regressors in this example, it is able to properly formulate the auto-regressive nature of electricity consumption. On the other hand, EBLR is able to incorporate this seasonality without having these explicit auto-regressive features.

\begin{table}[!h]
\centering
\caption{Three different probabilistic forecasting interval on a 14 hour Istanbul total electricity consumption. Each forecast horizon contains the median forecast (dark green line), the 50\% forecast interval (medium green fill), and 90\% forecast interval (light green fill).}
\label{tab:turkish_prob_res}
\begin{tabular}{ccccccc}
\hline
\multirow{2}{*}{Model} & \multicolumn{6}{c}{\textbf{WSPL($\rho$)}}                \\ \cline{2-7} 
        & 0.05   & 0.25   & 0.50   & 0.75   & 0.95   & Mean   \\ \hline
EBLR    & 0.0049 & 0.0149 & 0.0177 & 0.0130 & 0.0035 & 0.0108 \\
GBR     & 0.0089 & 0.0211 & 0.0226 & 0.0179 & 0.0079 & 0.0157 \\
RF      & 0.0036 & 0.0089 & 0.0105 & 0.0089 & 0.0039 & 0.0072 \\
Mean    & 0.0123 & 0.0395 & 0.0509 & 0.0353 & 0.0120 & 0.0300 \\
ARIMAX & 0.0075 & 0.0158 & 0.0148 & 0.0131 & 0.0055 & 0.0113 \\ \hline
\end{tabular}
\end{table}

\begin{figure}[!h]
    \centering
    \subfloat[EBLR \label{fig:turkish_prob_eblr}]{\includegraphics[width=0.5\textwidth]{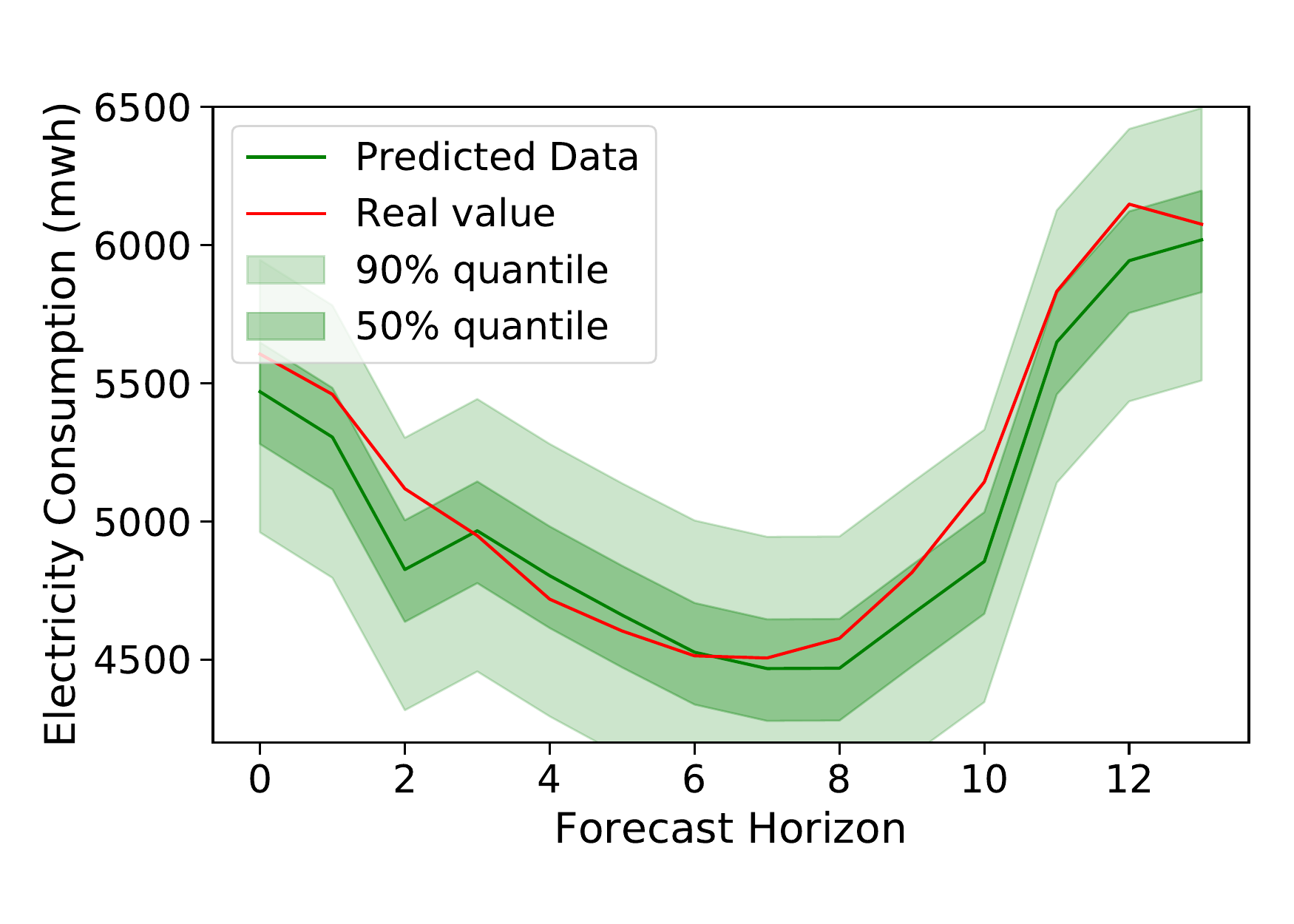}}
    \hfill
    \subfloat[GBR \label{fig:turkish_prob_gbr}]{\includegraphics[width=0.5\textwidth]{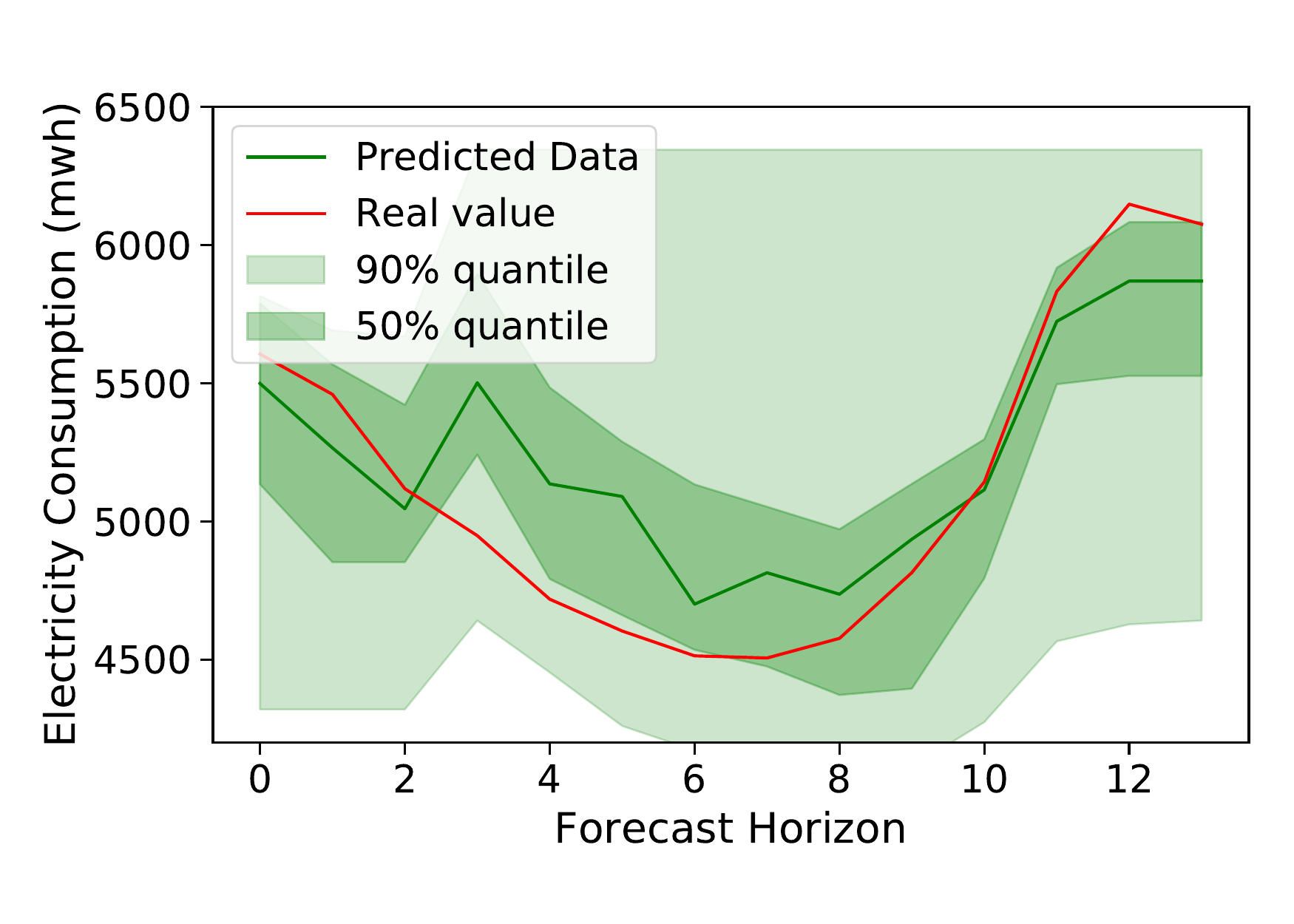}}\\
    \subfloat[ARIMAX \label{fig:turkish_prob_sarima}]{\includegraphics[width=0.5\textwidth]{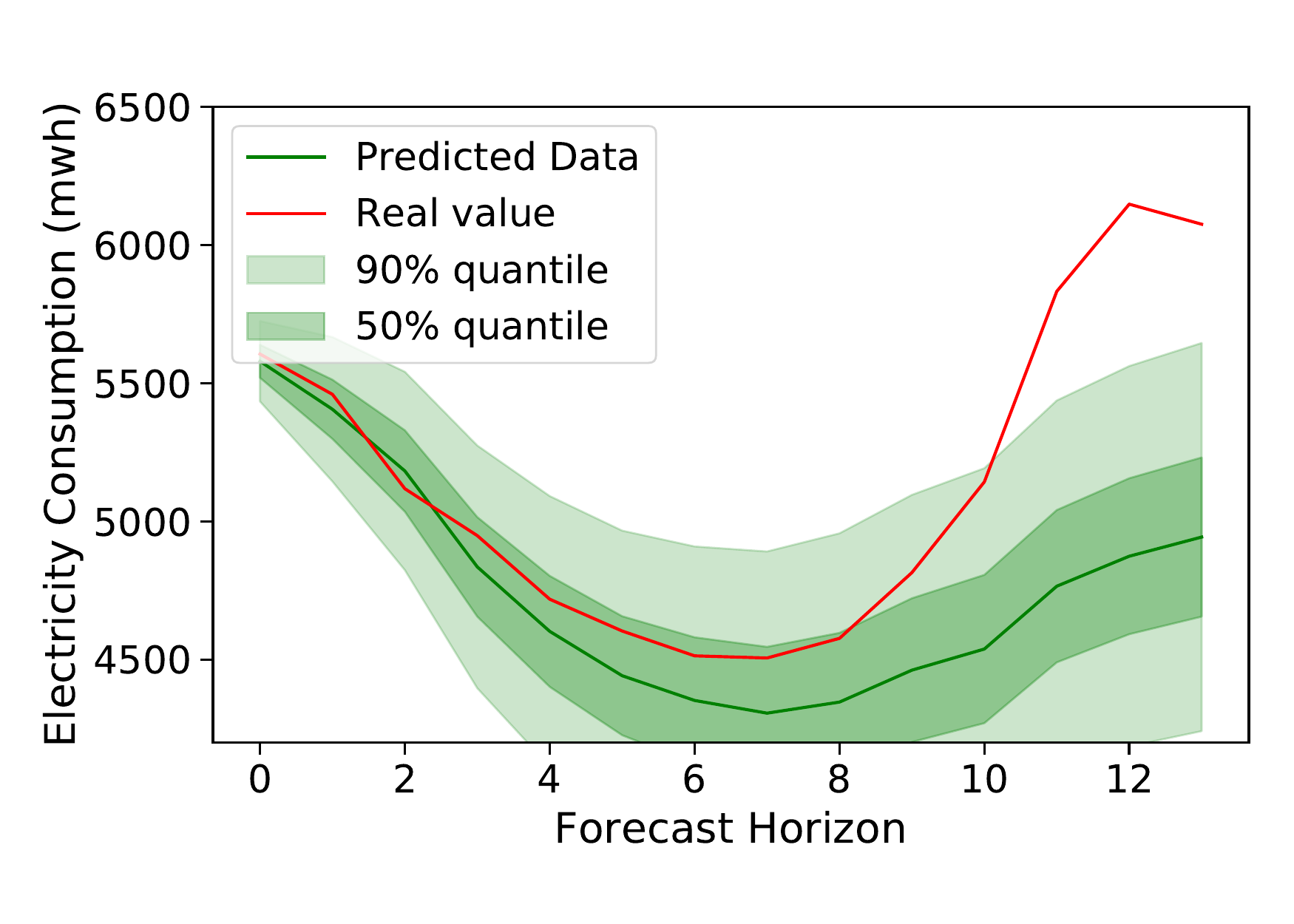}}
    \caption{Three different probabilistic forecasting interval results on a 14 hour electricity forecast horizon for electricity consumption in Istanbul. Each forecast horizon contains the median forecast (dark green line), the 50\% forecast interval (medium green fill), and 90\% forecast interval (light green fill).}
    \label{fig:probForecastsTurk}
\end{figure}

For posterity, EBLR was compared to a DeepAR model consisting of two LSTM layers of 40 cells each \cite{salinas2020deepar}. The comparison is realized on the Rossman dataset since deep learning methods require very large number of instances to properly function \cite{salinas2020deepar}.
%The only reasonable dataset to perform this comparison was the Rossmann data set. All other data sets are single time series, and the DeepAR package creators explicitly recommend that best practices are only to use DeepAR on \textit{``dataset(s) contains hundreds of related time series''} \footnote{https://docs.aws.amazon.com/sagemaker/latest/dg/deepar.html}. 
%They do this because DeepAR is designed to generate a global model from many local models. From the three datasets reported, only one of them fits this criterion. 
Although it has a highly complex structure,
%still, with all of DeepAR's overhead, it was 
DeepAR produces an NRMSE of 0.1495 which is only a marginal improvement over EBLR.
%only yields marginally better performance than EBLR. DeepAR produced an NRMSE of 0.1495, only a 2.1\% improvement on EBLR. 
In terms of probabilistic forecasts, the average WSPL across the considered quantiles for DeepAR is 0.0320. This result is worse than GBR's average WSPL, and marginally better than EBLRs. As well, all this overhead comes at the cost of interpretability since DeepAR is regarded as a black-box model.
% \begin{figure}[!htp]
%     \hspace*{-3cm} 
%     \centering
%     \includegraphics[scale=0.6]{images/turkish-interval-forecasts-full.pdf}
%     \caption{Probabilistic Turkish Examples}
%     \label{fig:turkish_prob}
% \end{figure}

% Once again, if we include the features from EBLR to SARIMAX, we are able to further improve SARIMAX by ##%. Even if there is a strong prediction model, EBLR is able to generate features to improve the model.

\subsection{Model Interpretability}
\label{subsec:feat_imp}
%Sections \ref{sec:point_res} and \ref{sec:prob_res} illustrate the prediction performance of EBLR compared to the other methods on three datasets.
This section focuses on the explainability by illustrating the interpretable nonlinear features generated by EBLR for these datasets. It is illustrated in Section~\ref{sec:illustration} that EBLR is capable of discovering the interaction effects. For example, in the process of learning the underlying model of the synthetic data set, the following features are learned:
\begin{enumerate}
    \item Is the day a weekend with a promotion?
    \item Is the day a weekend without a promotion?
    \item Is the day a weekday with a promotion
    \item Is the day a weekday without a promotion?
\end{enumerate}
%The contributions of this work are two fold. As seen above, EBLR is able to generate predictions that are comparable to classic ensemble methods for a variety of datasets. Here, we focus on the second contribution, the interpretability.

%During EBLR's learning process, important non-linear features are learned. These features can be feed to other more complex models. 

This aligns with the synthesized features. First, EBLR learns how weekends deal with promotions, and then it determines how weekdays deal with promotions. This can clearly be seen in Figure~\ref{fig:eblr_learning}, where the strong weekend features are learned then the weekday features are learned. As well, after these four features are learned, the model is terminated since the regression tree can not find a split.

Similarly, in the Rossmann data set a key learned feature is if the date is a Monday, without any promotions, and no school holiday. This feature contributes negatively to the prediction which is intuitive to understand. People typically do not shop on Mondays unless there is a special event. Other contributing features are tied relations between the month of the year, promotional activity, and school holidays. This is easily extracted through EBLR, which provides valuable data insights.

\begin{figure}[!h]
    \centering
    \includegraphics[width=0.5\textwidth]{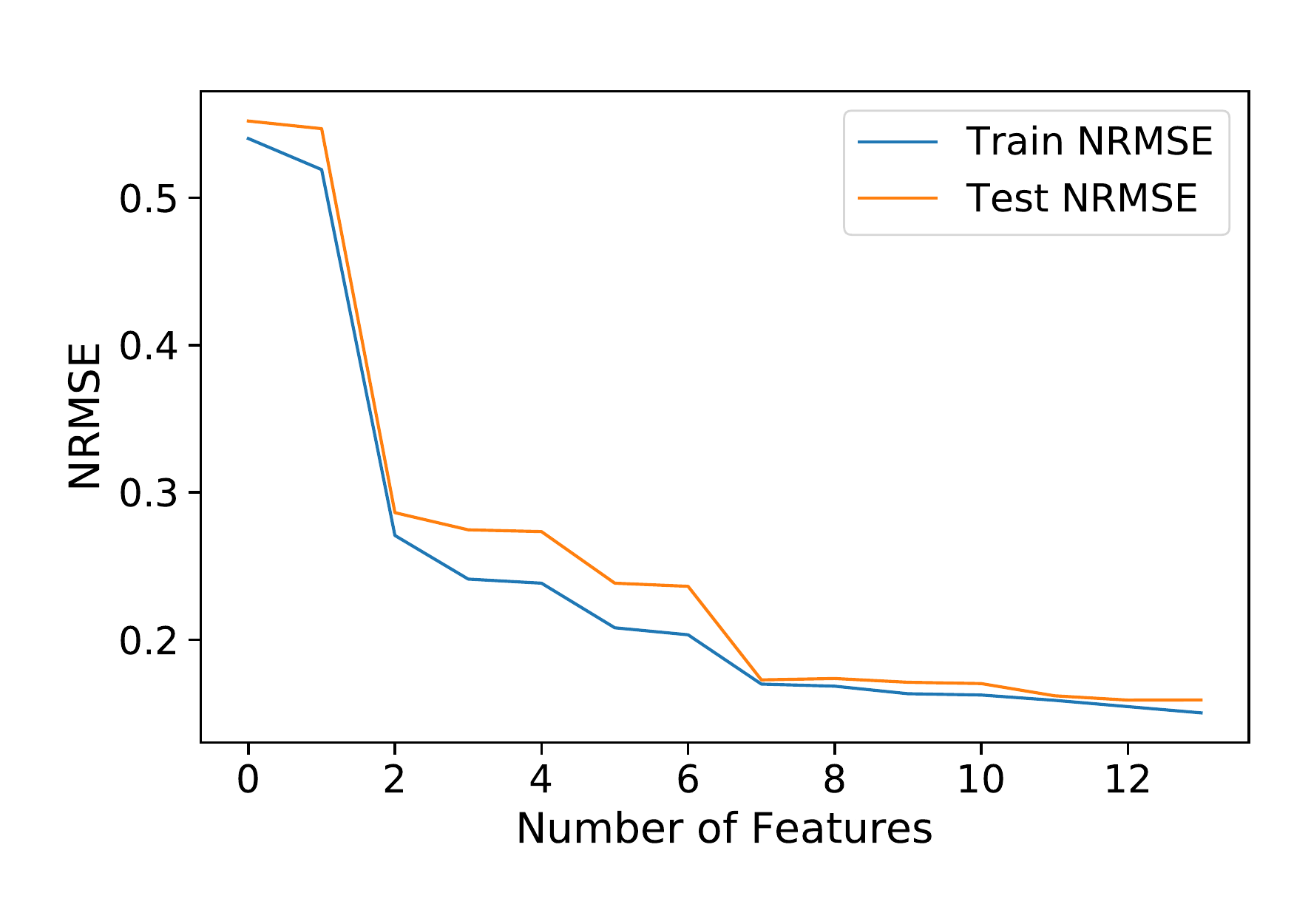}
    \caption{EBLR's improvement in NRMSE for a sample Rossmann sales store, as more features are generated and added to the base linear learner.}
    \label{fig:rossman_learning}
\end{figure}

The progression of EBLR's learning process can be seen through plotting the NRMSE against the number of features. A sample learning curve has been plotted for a particular Rossman store in Figure~\ref{fig:rossman_learning}.

Feature importance scores are generated based on these rules, combined with the learning curve. For each rule that is created, there is a change in the residual error, $\Delta e$. Earlier learned features decrease the residual error tremendously, whereas fine-tuning features are deemed not as important. This error is allocated to all boolean decisions in the feature. For example, when EBLR first learns the feature (Is Weekend, Yes),  ${\Delta e_1}$ is assigned to each feature for $isWeekend$ and $isPromo$. This is done for each feature generated, and then all the scores are added together and normalized. For example, in the synthetic dataset in Section \ref{sec:illustration}, every feature generated consists of a combination of $isPromotion$ and $isWeekend$. Since these features exist in all the rules, as well as the only features that EBLR utilizes, they both receive an equal feature importance score of 0.5.

\begin{figure}[!h]
    \centering
    \includegraphics[width=0.5\textwidth]{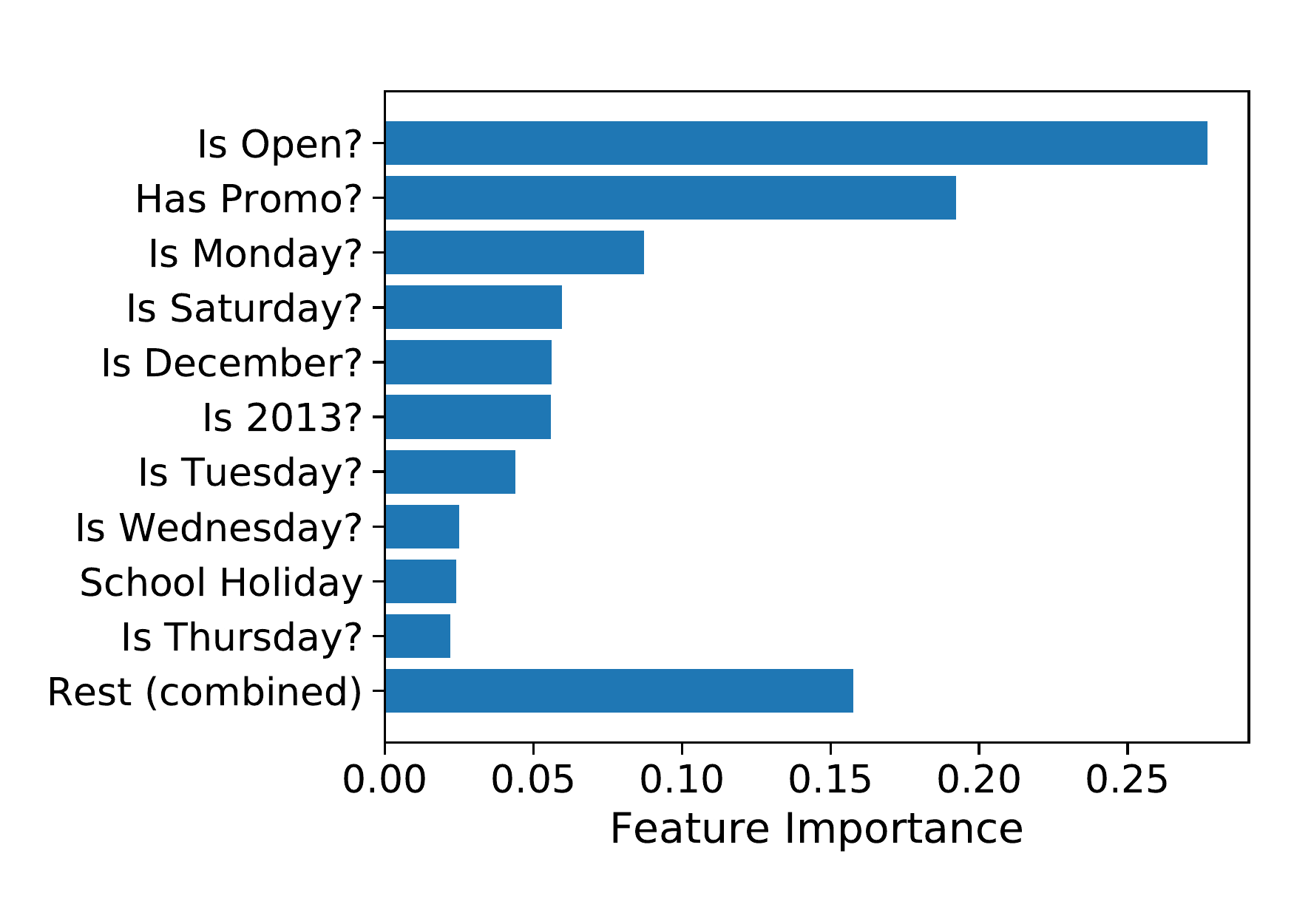}
    \caption{The top 10 important features in all of the Rossmann stores, organized from most important to least.  They are equally weighted across all stores to generate all feature importance's in the Rossmann dataset.}
    \label{fig:rossman_featimps}
\end{figure}

In Figure \ref{fig:rossman_learning}, there are initially a few features that drastically improve the model. Then, the model learns more complex non-linear features to continue learning. At each iteration, there is a change in how the underlying baseline linear model performs. By utilizing the proposed feature importance scoring technique, the most contributing features in the generated rules are extracted. A plot of the feature importance scores of the top 10 most important features in the Rossmann dataset in Figure \ref{fig:rossman_featimps}.

For the Turkish electricity data set, there is a wider spread of feature importance as observed in Figure~\ref{fig:turkish_featimps}. The two most important features are the daily high and lower temperatures, followed by information about the hour of the day and if the day was a Sunday. Together, these feature importance scores imply that electricity usage is highly dependent on temperature and the hour of the day. EBLR cares about knowing if the day is a Sunday or not, which means Sundays behave differently than the remaining days of the week. As well, there is an even distribution of feature usage in the Turkish electricity dataset, compared to a few key features in the Rossmann dataset.

\begin{figure}[!h]
    \centering
    \subfloat[Sample Learning Curve \label{fig:turkish_featimps}]{\includegraphics[width=0.5\textwidth]{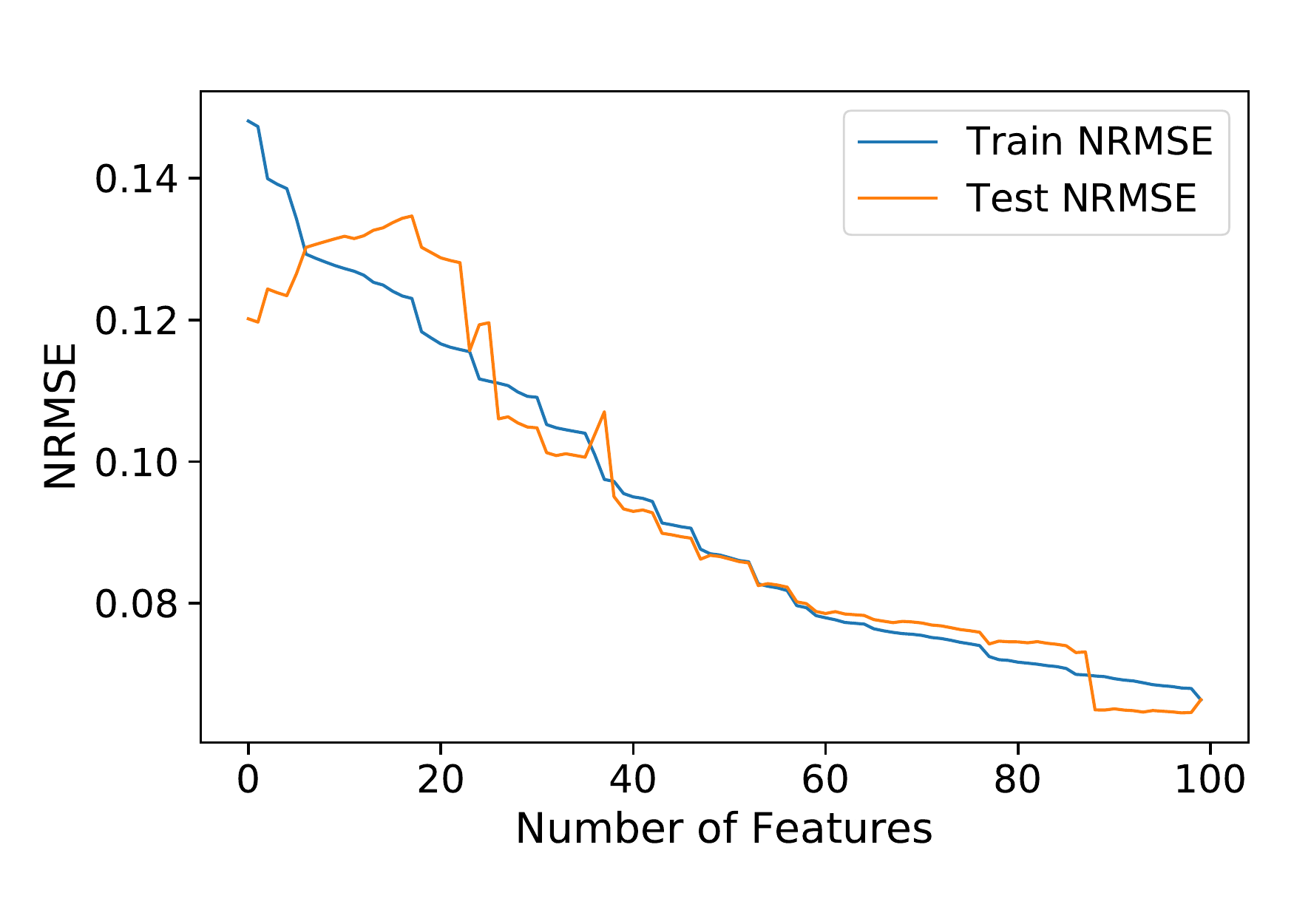}}
    \hfill
    \subfloat[Top 10 Feature Importance \label{fig:turkish_featimps}]{\includegraphics[width=0.5\textwidth]{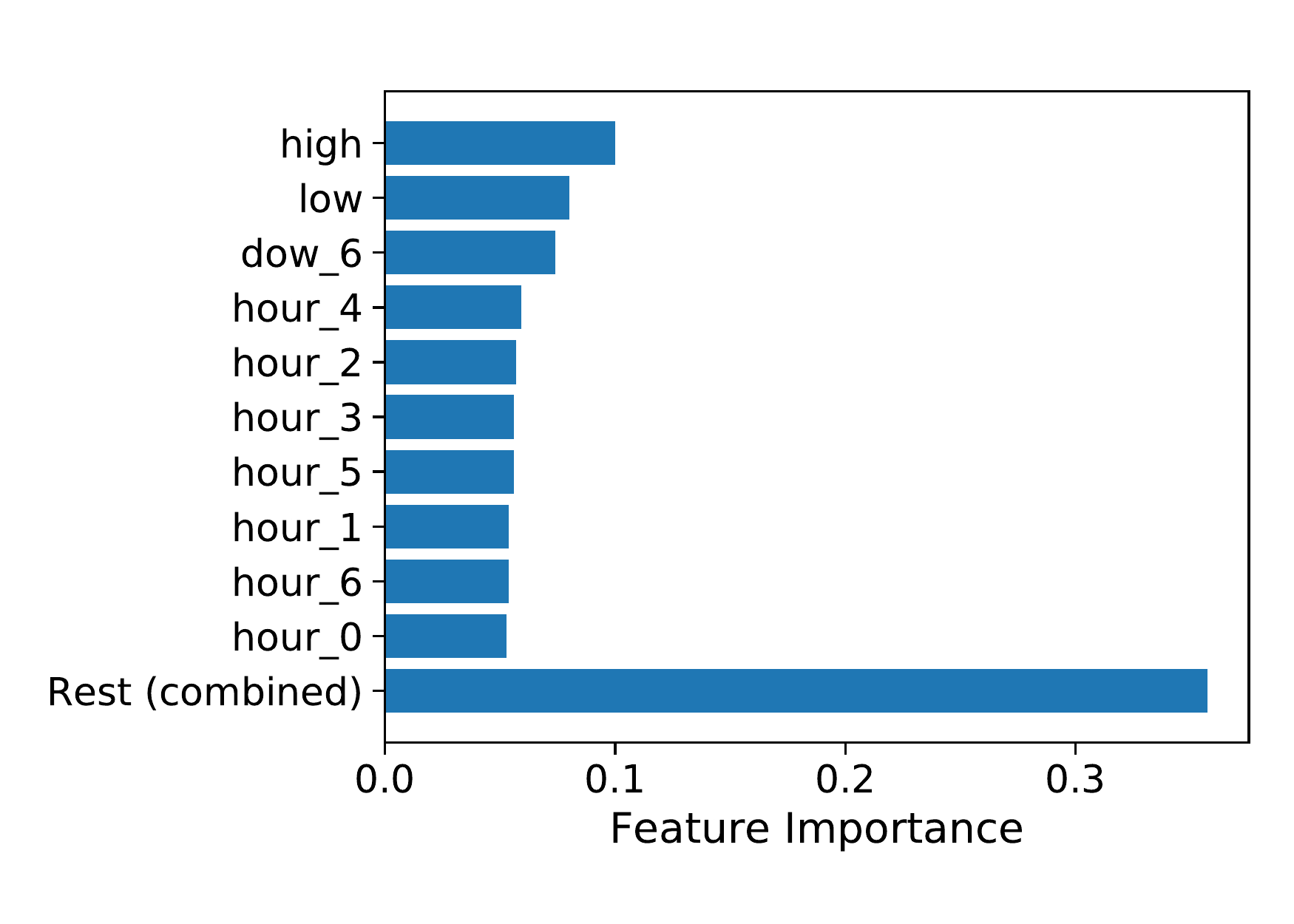}}
    \caption{(Left) The NRMSE learning curve compared to the number of features in a the Turkish electricity dataset. (Right) The top 10 most importance features in the Turkish electricity consumption dataset, organized from most important to least.}
    \label{fig:probForecastsTurk}
\end{figure}

%%%%%%%%%%%%%%%%%%%%%%%%%%%%%%%%%%%%%
\section{Conclusion}\label{sec:conclusion}
%%%%%%%%%%%%%%%%%%%%%%%%%%%%%%%%%%%%%
Through the probabilistic and point forecasting experimentation, it is evident that EBLR is a strong boosting algorithm. While there was no clear separation between the EBLR, GBR, and RF, it is clear that EBLR is much simpler than the other two ensemble methods. EBLR consists of simple binary features, compared to the complexity of storing numerous decision trees. On top of this, the information lost by only storing these relevant binary features is minimal. 

EBLR generates simple binary features through a two-step process. First, a simple baseline model is fit to a training dataset. The residuals are extracted and passed into a feature generating decision tree. This decision tree extracts the largest source of error in the form of an interpretable feature which is passed back into the baseline learner. This process is repeated until a stopping condition is met. By learning in this manner, EBLR has inherent interpretability baked into itself.

To extend EBLR to generate probabilistic forecasts, an empirical distribution is generated from EBLR's training residuals. Quantiles are selected from this distribution, and used in making prediction intervals. While this was able to yield strong results in the three provided experiments, future work should focus on different ways to generate these prediction intervals. By generating prediction intervals in this manner, there are constant prediction intervals across all points. In reality, some specific points are more difficult to predict than others. Some potential research paths include using a quantile base linear learner or extracting more information from the feature generating decision trees.

\bibliographystyle{elsarticle-harv}
\bibliography{eblr_refs}

\end{document}